# A Transformer-based survival model for prediction of all-cause mortality in heart failure patients: a multi-cohort study

Word count: 3,845

Shishir Rao, DPhil[1], Nouman Ahmed, MSc[1], Gholamreza Salimi-Khorshidi, DPhil[1], Christopher Yau, DPhil[1,2], Huimin Su, MPhil[3], Nathalie Conrad, DPhil[1,3], Folkert W Asselbergs, MD PhD[2,4,5], Mark Woodward, PhD[6,7], Rod Jackson, PhD[8], John GF Cleland, MD[9], Kazem Rahimi, DM[1*]

[1] Nuffield Department of Women's & Reproductive Health, University of Oxford, Oxford, United Kingdom
[2] Health Data Research UK, London, UK
[3] Department of Cardiovascular Sciences, Katholieke Universiteit Leuven, Belgium
[4] Amsterdam University Medical Centers, Department of Cardiology, University of Amsterdam, Amsterdam, The Netherlands.
[5] Institute of Health Informatics, University College London, London, UK.
[6] The George Institute for Global Health, University of New South Wales, Newtown, NSW, Australia
[7] The George Institute for Global Health, Imperial College London, London, United Kingdom
[8] School of Population Health, Faculty of Medical and Health Sciences, University of Auckland, Auckland, New Zealand
[9] British Heart Foundation Centre of Research Excellence, School of Cardiovascular and Metabolic Health, University of Glasgow, Glasgow, UK

*Corresponding author: Kazem Rahimi, Deep Medicine, Oxford Martin School, University of Oxford, Oxford, United Kingdom; email address: kazem.rahimi@wrh.ox.ac.uk; Tel: +44 (0) 1865 617200; Fax: +44 (0) 1865 617202; Postal code and address: Nuffield Department of Women's and Reproductive Health, University of Oxford, Level 3, Women's Centre, John Radcliffe Hospital, Oxford, OX3 9DU, United Kingdom

## Abstract

**Background**: Accurately predicting clinical outcomes in patients with heart failure (HF) can help inform patient management and service audits. Current risk scores rely on specialised tests (e.g., ejection fraction), but offer modest discrimination in large part as they fail to capture the evolving multi-factorial risk profiles of HF patients.

**Methods**: Using routine UK electronic health records (EHR), we developed and validated the Transformer-based Risk assessment survival model (TRisk), an artificial intelligence model that predicts 36-month mortality in HF patients by analysing temporal patient journeys – including diagnoses, medications, and procedures – up till baseline. A cohort of 403,534 HF patients aged 40 to 90 years was identified from 1,418 English general practices (1,063 for derivation and the remaining 355 for external validation). TRisk was compared with the Meta-Analysis Global Group in Chronic HF model adapted for use with routine linked EHR (named as MAGGIC-EHR), with analyses across age groups, sex, and baseline characteristics. A second validation of TRisk using transfer learning was conducted on 21,767 HF patients from USA hospital admission data. Lastly, explainability analysis examined which features TRisk prioritised for prognostication in each cohort.

**Findings**: In the UK cohort, with median follow-up of 9 months (interquartile interval: [2, 29]), TRisk demonstrated a concordance index (C-index) of 0.845 (95% confidence interval: [0.841, 0.849]), outperforming MAGGIC-EHR with a C-index of 0.728 (0.723, 0.733) for predicting 36-month all-cause mortality. In subgroup analyses, TRisk demonstrated less variability in predictive performance by sex, age, and baseline characteristics compared to MAGGIC-EHR, suggesting a less biased modelling. Adapting TRisk to USA data using transfer learning yielded a C-index of 0.802 (0.789, 0.816) for predicting all-cause mortality. Explainability analyses showed TRisk captured established risk factors while also identifying several underappreciated factors, notably cancers and hepatic failure, that were consistently important across both cohorts. Additionally, the prognostic utility remained strong for cancers occurring even a decade before baseline, while other factors weakened with time.

**Interpretation**: TRisk demonstrated well-calibrated, accurate prediction of mortality across both UK and USA healthcare settings using routinely collected EHR. Explainability analyses identified several risk factors not included in previous expert-driven models, underscoring the value of tracking longitudinal and evolving health profiles. By capturing complex patient journeys through routine EHR, the model would better stratify HF patients and guide management.

**Funding:** Horizon Europe

# Introduction

Heart failure (HF) is a complex clinical syndrome with a highly variable prognosis.[1] Risk prediction of outcomes over a few days or over many decades is relatively accurate in HF but predicting medium-term outcomes over a few months to years is often difficult. Prediction in this time range is, however, important for initiating interventions, effectively communicating with patients and families, and auditing the quality of care across organisations.[2–4]

While risk assessment approaches ranging from simpler single markers (e.g., left ventricular ejection fraction [LVEF]) to more comprehensive models like the Meta-Analysis Global Group in Chronic (MAGGIC) HF have been useful, limitations remain.[3] Many rely on data from specialised tests that are resource-intensive to obtain and are typically gathered after clinical suspicion of HF deterioration has been raised. Moreover, these models offer relatively modest discrimination (i.e., <0.8 concordance index) for select outcomes with poor positive predictive value (PPV) and sensitivity.[3,5–9] While they focus primarily on including predictors of cardiovascular function, current models fail to capture the complex, evolving nature of multi-factorial risk, overlooking important comorbidities and interventions that influence outcomes. In fact, complex multimorbidity, rather than HF itself, contributes to over 40% of mortality in HF patients, highlighting a significant gap in current risk assessment approaches.[10–12] Given these limitations, current models have consequently seen limited clinical adoption, with HF guidelines emphasising the need for more robust models with transparent reporting of performance metrics.[7,8,13]

The widespread adoption of electronic health record (EHR) systems offers the potential for developing more sophisticated risk assessment models that capture dynamic, multi-factorial patient risk profiles using routine clinical data. Emerging as an efficient approach for harnessing such large-scale EHR in predictive modelling, the Bidirectional EHR Transformer (BEHRT) and its survival modelling variant, the Transformer-based Risk assessment survival model (TRisk) represent a breakthrough in patient health modelling.[14,15] These AI models capture the complete patient journey by processing various medical data types—diagnoses, medications, procedures, and test results—as recorded over time, advancing prognostication research across multiple fields, including cardiology, oncology, and pulmonology.[14–17] Furthermore, as TRisk has been designed to operate on data standardised according to internationally accepted dictionaries of representing clinical data (e.g., SNOMED), the model demonstrates promise for application on global data. Motivated by these factors, we developed and validated TRisk for prediction of all-cause mortality in HF patients using United Kingdom (UK) and United States of America (USA) EHR datasets.

# Methods

## Overview

Using a UK cohort of patients with HF, we derived TRisk and a version of the MAGGIC Cox proportional hazards (CPH) model adapted for use on primary and secondary care EHR (henceforth, termed MAGGIC-EHR).[3,15] All models were externally validated in a separate cohort of UK patients with HF; additional external validation of TRisk for all-cause mortality prediction was conducted in a cohort of patients with HF from the USA. To understand decision-making processes of the "black-box" TRisk model, we analysed the input EHR encounters the model found as important for conducting prediction across both cohorts.

### Data sources
For the UK data study, we used English data provided by the Clinical Practice Research Datalink (CPRD) Aurum dataset. CPRD includes detailed patients' records including demographics, diagnoses, prescribed treatments, and health-related lifestyle variables collected from participating general practices (GPs) across UK, covering approximately 20% of the UK population.[18] For our English GP data, CPRD provided linkage to Hospital Episode Statistics (HES) and Office for National Statistics (ONS) for data on hospital visits and mortality reports, respectively.[18]

For the USA data study, we used the Medical Information Mart for Intensive Care IV (MIMIC-IV) hospital admissions dataset covering admissions at Beth Israel Deaconess Medical Center in Massachusetts, USA, from 2008 to 2019.[19] This dataset contains health records from hospital admissions, including patient demographics, diagnoses, vital signs, laboratory test results, medications, and procedures. Additionally, the dataset provides linkages to the Center's outpatient records and Massachusetts State Registry of Vital Records and Statistics (RVRS) for data on outpatient visits and mortality records respectively.[18]

### Primary outcome
The primary outcome of interest was all-cause mortality with date of outcome captured using linked ONS and RVRS mortality records for UK and USA data respectively.

### Validation strategy
We included data from 1,418 contributing GP practices from England. We randomly assigned three quarters (i.e., 1,063) of practices to the derivation dataset and the rest (i.e., 355) were allocated to a dataset for external validation (details in **Supplementary Methods: Clarification on CPRD validation study**).

### Cohort selection, index date, and follow-up period definition
We identified an open cohort of individuals between the ages of 40 and 90 years with first HF diagnosis in the study period. The study period for each patient was defined as starting from the latest of the following: January 1st 2005, patient's 40th birthday, and 12 months after GP registration. It ended at the earliest of the following: date of outcome of interest (i.e., mortality), last collection of data from practice, patient's last date in practice, or December 31st 2019. We excluded individuals who had an HF diagnosis before the study period. HF was identified by previously published phenotyping dictionaries for CPRD Aurum.[20]

For each HF patient selected, the index date (i.e., baseline), was randomly selected from the period between first HF diagnosis and end of study-period. This approach simultaneously ensures capture across the spectrum of calendar years and ages, and better reflects the reality of clinical practice, where patients are assessed at various stages of their care journey.[21] Patients were censored at study period end date.

### AI and statistical modelling
TRisk is a deep learning model that adopts the survival ordinary differential equations network-based (SODEN) framework for survival prediction with additional regularisation procedures for better calibrated predictions compared to simpler AI models (**Supplementary Methods: EHR pre-processing and AI modelling**).[22,23] For input, TRisk considers a patient's medical history up to baseline as a sequence of timestamped records.[14,15] The model incorporates all records from the following modalities in linked primary and secondary care records: 2,639 distinct diagnosis codes, 633 medication codes, and 852 procedure codes. Relative ordering and patient age at each encounter (e.g., diagnosis) are also provided to the model, thereby providing rich longitudinal annotations to the sequential stream of encounters in medical history up to index date (**Figure S1**). The input provided to TRisk is as captured in routine EHR without imputation of missing values (**Supplementary Methods: EHR pre-processing and AI modelling**; model parameters in **Table S1**).

The MAGGIC-EHR model was adapted as a benchmark model for application on primary and secondary care EHR data.[3] All variables except for LVEF, which is not typically available in large-scale population-based EHR, were extracted from patient records and used in MAGGIC-EHR modelling. In lieu of the LVEF measurement, a categorical variable indicating HF subtype (i.e., reduced, preserved, and unknown LVEF), derived using published phenotyping codes, was incorporated into the model as a practical substitute.[24] Extraction and transformation of MAGGIC-EHR model predictors followed roughly the specifications of previously published work.[3] To address data missingness, imputation was separately carried out on derivation and validation datasets.[25] More information on predictor extraction, imputation for benchmark modelling, and implementation can be found in **Supplementary Methods: Implementation of MAGGIC-EHR models**.

Performance analysis on UK validation data

We evaluated models' performance on the UK validation dataset using multiple metrics. For discrimination, we assessed the concordance index (C-index) and area under the precision-recall curve (AUPRC). Model calibration was assessed graphically, by comparing the agreement between the observed and predicted risk using a smoothed calibration curve fit using restricted cubic splines with three knots, and quantitatively by calculating the integrated calibration index (ICI) metric.[26] Decision curve analysis (accounting for censoring) assessed the balance between true positives (correctly identified mortality) and false positives across the risk spectrum.[27] For all analyses, individual patient risk scores were derived by estimating survival probabilities over a 48-month follow-up period, focusing on the 36-month timepoint.

Additionally, we conducted discrimination analyses in subgroups defined by sex, age, baseline systolic blood pressure, HF subtype (i.e., preserved or reduced LVEF as identified by algorithm introduced previously[24]), baseline disease status, and baseline medication use.

To ensure fairer comparison with TRisk, we implemented MAGGIC-EHR with an extended predictor set (denoted henceforth as MAGGIC-EHR+) that included additional established risk factors: serum sodium concentration as well as history of atrial fibrillation, stroke, myocardial infarction, percutaneous coronary intervention, and coronary artery bypass grafting procedure.[2]

Calibration and net benefit analyses for 12-month prediction were also conducted for all models to assess model performance at different follow-up durations.[2]

Furthermore, we conducted an impact analysis to evaluate true positive and negative capture at the 50% decision threshold of the TRisk and MAGGIC-EHR models on UK validation data in line with research recommendations by HF clinical guidelines.[13]

Analyses of USA data

We conducted additional external validation of TRisk for 36-month all-cause mortality prediction in the MIMIC-IV USA dataset as an entirely independent cohort. Following similar cohort selection used on CPRD, we randomly allocated 60% of the identified HF patient cohort to the validation dataset while the remaining 40% formed the fine-tuning dataset (details in **Supplementary Methods: Clarification on MIMIC-IV validation study**).

For analysis on the MIMIC-IV dataset, we utilised transfer learning to transport knowledge learned by the TRisk model from the UK data setting to the USA data setting.[28] Specifically, we utilised the TRisk model trained on CPRD derivation dataset and then fine-tuned (i.e., conducted transfer learning) on the MIMIC-IV fine-tuning dataset. We evaluated the model on the MIMIC-IV validation dataset and conducted discrimination, calibration, and decision curve analysis.

For supplementary analysis, to compare against the transfer learning approach, two additional variants of TRisk were evaluated on MIMIC-IV validation dataset: one model solely trained on CPRD data (i.e., conventional external validation) and another solely trained on the MIMIC-IV fine-tuning dataset (i.e., internal validation). Details on all modelling can be found in **Supplementary Methods: Clarification on modelling on MIMIC-IV dataset**. For all models, we conducted discrimination, calibration, and decision curve analyses. Lastly, all analyses were replicated for 12-month mortality prediction.

### Explainability analyses of TRisk
To understand TRisk's decision-making processes, we used the integrated gradients method to analyse which medical history encounters most influenced risk score calculation.[29] This method measures how each encounter (e.g., diagnosis, prescription, procedure) in a patient's medical history contributes to the risk score; to capture each encounter's population-level importance, we then averaged these contribution scores across the cohort.

Our population analysis consisted of two stages: validating TRisk's ability to detect established clinical risk factors[2,3], then determining which encounters it independently identified as most significant. We analysed these patterns across sex, age, and time from first encounter to baseline in the UK validation cohort. To ensure robustness of our explainability discoveries, we conducted parallel validation on the USA validation cohort (details in **Supplementary Methods: Explainability analyses**).

### Analyses of secondary outcomes on UK data
Finally, we conducted supplementary analyses to assess the utility of TRisk for prediction of outcomes other than all-cause mortality. Leveraging comprehensive follow-up data offered by CPRD in tandem with validated phenotyping for outcome identification, we conducted 36-month prediction of three additional outcomes identified through primary care, secondary care, and mortality records in UK EHR: (1) fatal and non-fatal cardiovascular event prediction (i.e., composite of ischaemic heart disease, myocardial infarction, transient ischaemic attack, and stroke), (2) cardiovascular-related mortality prediction (identified solely in mortality records), and (3) renal outcomes prediction (i.e., composite of chronic kidney and end-stage renal diseases). Outcomes were identified using both previously published and our own curated code dictionaries.[20,30] Similar to all-cause mortality prediction analysis, we evaluated all models on the UK validation data and discrimination, calibration and net benefit for outcome investigations were assessed; analyses were replicated for 12-month prediction of all outcomes.[2]

### Code availability
All software for conducted analyses and phenotyping dictionaries for identifying outcomes have been uploaded to the TRisk code repository (https://github.com/deepmedicine/TRisk).

### Role of the funding source
The funders of the study had no role in the study design, data collection, data analysis, data interpretation, writing of the report, or the decision to submit the report for this research.

## Results
### Analyses of UK Data
From UK data, 403,534 patients (**Figure S2**) were selected for analysis (99,382 patients in validation) with median follow-up of 9 (interquartile interval [IQI]: [2, 29]) months. The median age was 79 (IQI: [70, 85]) years and approximately, one-third of the cohort had a history of diabetes, myocardial infarction, or were prescribed beta-blockers at baseline (**Table 1**). Approximately, 44% of patients died in four years following baseline.

*Table 1. Population characteristics for derivation and validation datasets in UK data study at baseline*

|  | Derivation (304,152 patients) | Validation (99,382 patients) |
|---|---|---|
| Women (%) | 139,762 (46.0) | 45,444 (45.7) |
| Median age (years) (IQI) | 79 (70, 85) | 79 (70, 85) |
| Smoking status[†] | | |
| No (%) | 133,284 (43.8) | 43,605 (43.9) |
| Ex (%) | 131,651 (43.3) | 42,927 (43.2) |
| Yes (%) | 39,217 (12.9) | 12,850 (12.9) |
| Median BMI (kg/m$^2$) (IQI)[†] | 27.5 (24.0, 31.8) | 27.5 (24.0, 31.7) |
| Median SBP (mmHg) (IQI)[†] | 134.0 (124.9, 143.1) | 133.9 (124.9, 143.1) |
| Median sodium (mmol/L) (IQI)[†] | 139.6 (137.5, 141.3) | 139.5 (137.5, 141.2) |
| Median creatinine (μmol/L) (IQI)[†] | 91.3 (76.0, 113.4) | 91.8 (76.0, 113.8) |
| HF subtype (by LVEF status) | | |
| Unknown (%) | 225,948 (74.3) | 73,564 (74.1) |
| HFpEF (%) | 12,997 (4.3) | 4,350 (4.4) |
| HFrEF (%) | 65,207 (21.4) | 21,468 (21.6) |
| <18 months since incident HF (%) | 208,977 (68.7) | 68,279 (68.7) |
| NYHA Classification[†] | | |
| I (%) | 76,044 (25.0) | 24,700 (24.9) |
| II (%) | 143,318 (47.1) | 46,661 (47.0) |
| III (%) | 76,591 (25.2) | 25,264 (25.4) |
| IV (%) | 8,199 (2.7) | 2,757 (2.8) |
| Disease history | | |
| Diabetes (%) | 91,740 (30.2) | 29,659 (29.8) |
| COPD (%) | 81,100 (26.7) | 27,099 (27.3) |
| AF (%) | 146,493 (48.2) | 48,033 (48.3) |
| Stroke (%) | 70,675 (23.2) | 22,604 (22.7) |
| MI (%) | 86,429 (28.4) | 28,500 (28.7) |
| Prescription history | | |
| Beta-blockers (%) | 102,741 (33.8) | 34,258 (34.5) |
| ACE-Is/ARBs (%) | 19,5311 (64.2) | 63,718 (64.1) |
| Procedure history | | |
| PCI (%) | 33,726 (11.1) | 11,059 (11.1) |
| CABG (%) | 18,190 (6.0) | 6,078 (6.1) |

%: percent; IQI: interquartile interval (25$^{th}$ and 75$^{th}$ percentiles); BMI: body mass index; SBP: systolic blood pressure; HFpEF: heart failure-preserved ejection fraction; HFrEF: heart failure-reduced ejection fraction; LVEF: left-ventricular ejection fraction; NYHA: New York Heart Association; COPD: Chronic obstructive pulmonary disease; AF: atrial fibrillation; MI: myocardial infarction; ACE-I: Angiotensin-converting-enzyme inhibitors; ARB: Angiotensin receptor blockers; PCI: Percutaneous coronary intervention; CABG: coronary artery bypass graft. [†]indicates missing variables; BMI (26.7% missingness), smoking status (14.0%), SBP (4.0%), creatinine (9.1%), NYHA (96.6%), sodium (9.7%).

TRisk achieved higher C-index (0.845; 95% CI [0.841, 0.849]) than MAGGIC-EHR (0.728; 95% CI [0.723, 0.733]) along with higher AUPRC **(Figures 1, S3; Table S2)**. The subgroup discrimination analyses yielded that the TRisk outperformed MAGGIC-EHR across key subgroups (**Figure 1; Table S3**) and importantly presented lesser deviation from the overall cohort discrimination across subgroup analyses as compared to MAGGIC-EHR. Both models were well calibrated as evidenced by general affinity of calibration curves to the reference curve (**Figures 2A, S4**) and low ICI estimates (**Table S4**). In terms of predictive distribution, TRisk demonstrated two peaks in the lowest and highest ends of the predicted risk spectrum implying more nuanced stratification of risk as compared to the MAGGIC-EHR model, which showed a unimodal distribution with a high concentration of predictions towards the middle (**Figures 2B, S4**). Decision curve analysis showed that TRisk provided significantly greater net benefit than other strategies (**Figure S4C**) across the spectrum of clinically relevant thresholds (up to ~0.6).

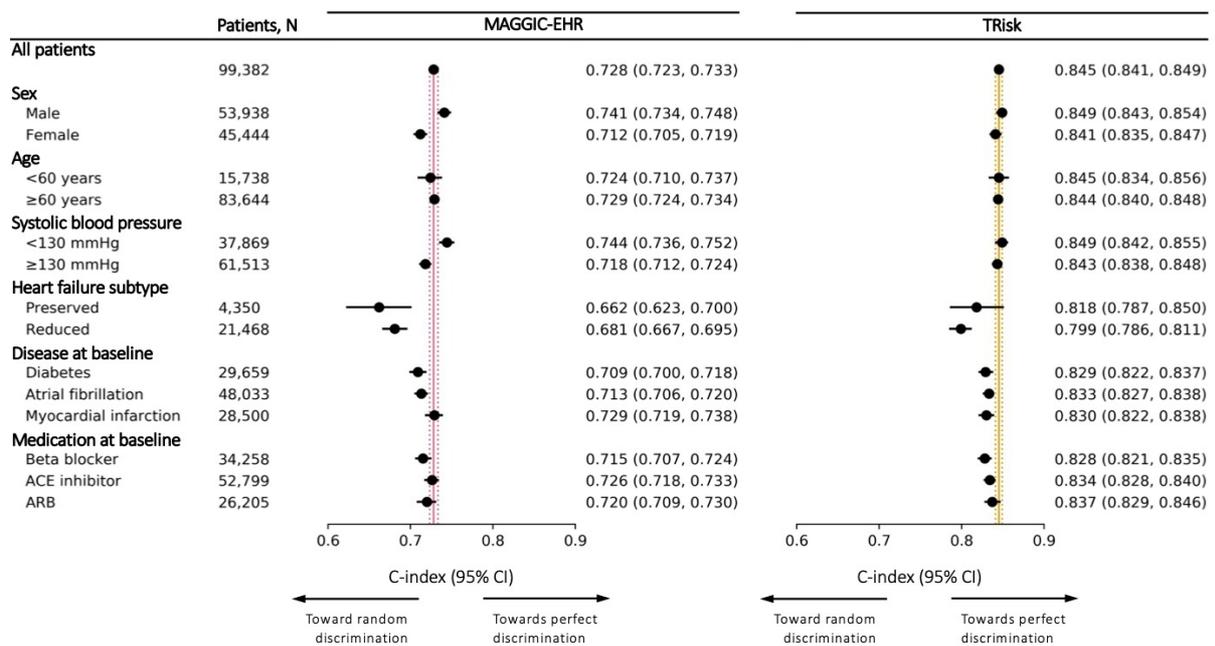

*Figure 1. Models' discrimination by concordance index (C-index) and associated 95% confidence intervals (CI) in overall cohort and subgroups for 36-month all-cause mortality risk prediction on UK validation data.* Maroon and gold lines represent C-index on "all patients" in the validation cohort for MAGGIC-EHR and TRisk respectively. These solid and dotted coloured lines represent mean and 95% confidence interval boundaries of the overall cohort C-index for each model. These lines are provided to visually demonstrate deviation in subgroup discrimination performance from overall cohort performance for each model. ACE: Angiotensin-converting enzyme; ARB: Angiotensin receptor blocker.

Investigating the performances of additional models, TRisk outperformed MAGGIC-EHR+, which performed similarly to MAGGIC-EHR (**Figures S3, S4; Tables S2-S4**). Similar findings were captured in analyses of 12-month mortality prediction (**Figure S5**; **Table S5**).

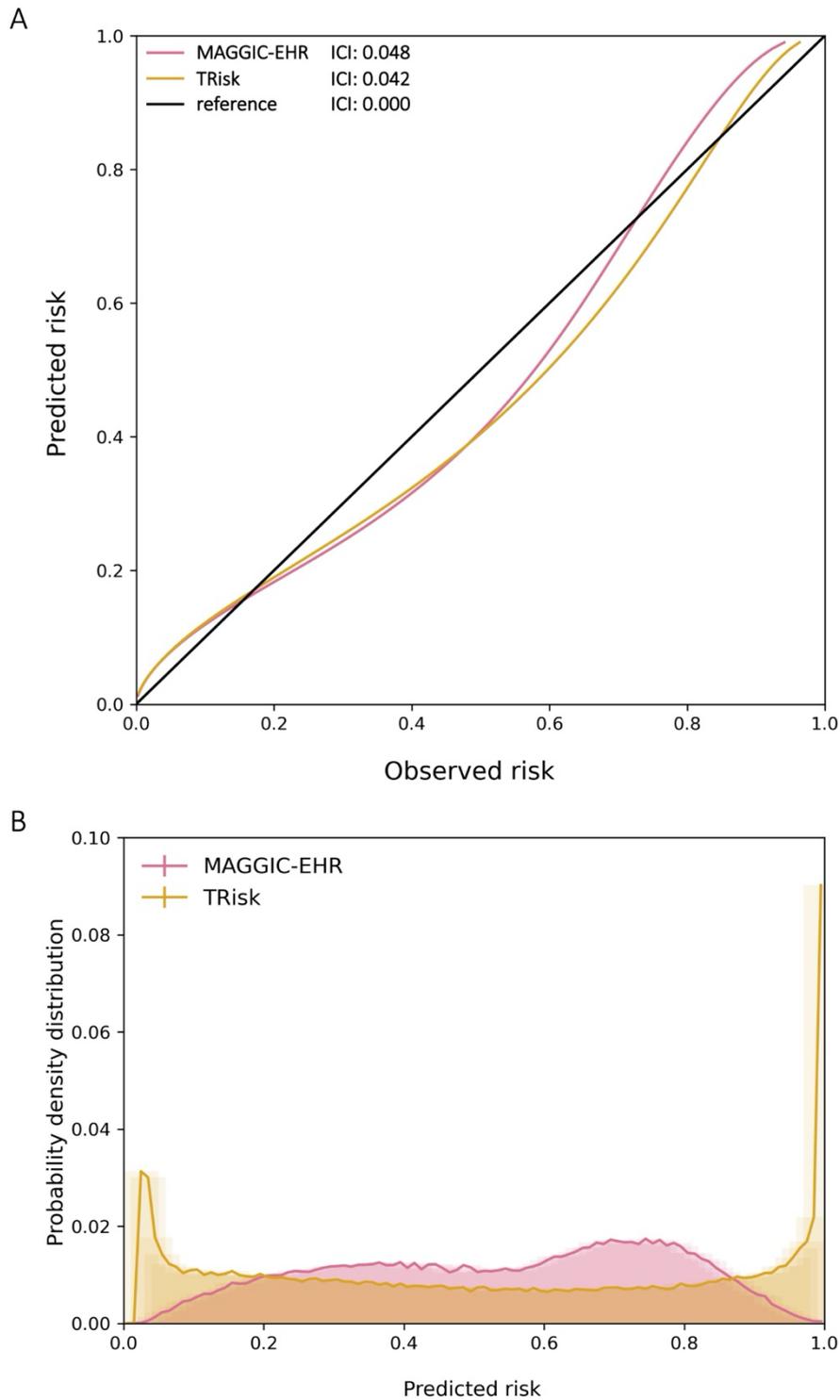

*Figure 2. Calibration curves and integrated calibration indices (ICI) and distribution of predicted risk of models for 36-month all-cause mortality risk prediction on UK validation data. Calibration curves with ICI (A) and distribution of predicted risk (B) are presented for all models. For (A) ICI, lower is better with reference (black line) presenting optimal ICI of 0.0.*

In impact analysis, TRisk reduced false positives and false negatives by 9% (699 patients) and 46% (8,544) respectively for 12-month prediction, and by 31% (7,693) and 24% (2,383) respectively for 36-month prediction, as compared to MAGGIC-EHR (**Figure 3; Table S6**).

Fewer false positives and false negatives directly translated into TRisk outperforming MAGGIC-EHR in both PPV and sensitivity, showing increases of 0.133 and 0.270 for 12-month prediction, and 0.105 and 0.059 for 36-month prediction, respectively (**Figure 3; Table S6**; further elaboration in **Supplementary Results: Impact analyses on CPRD validation dataset**).

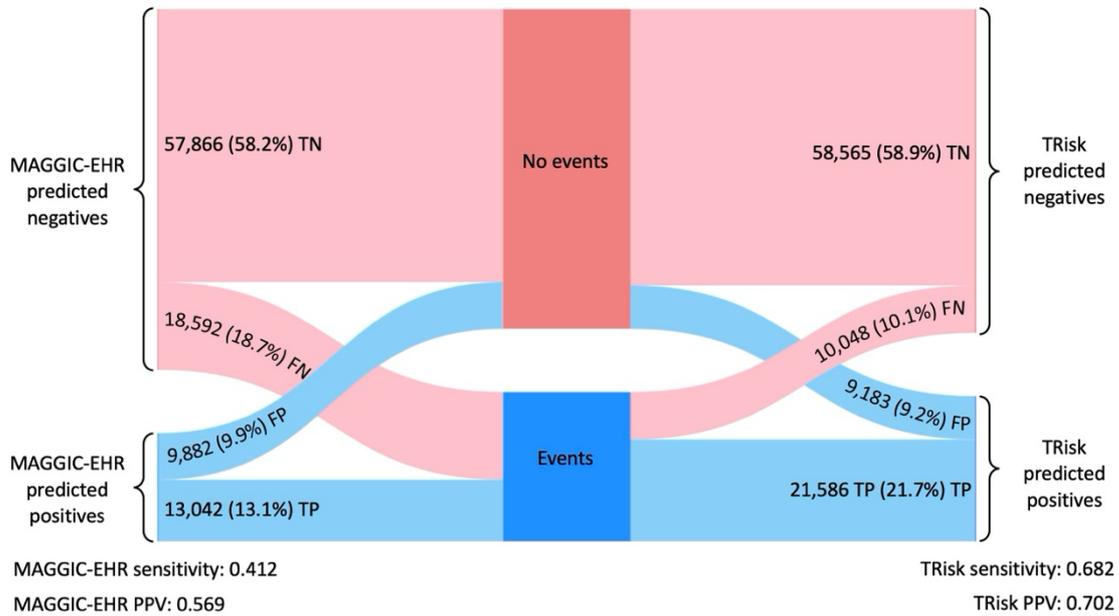

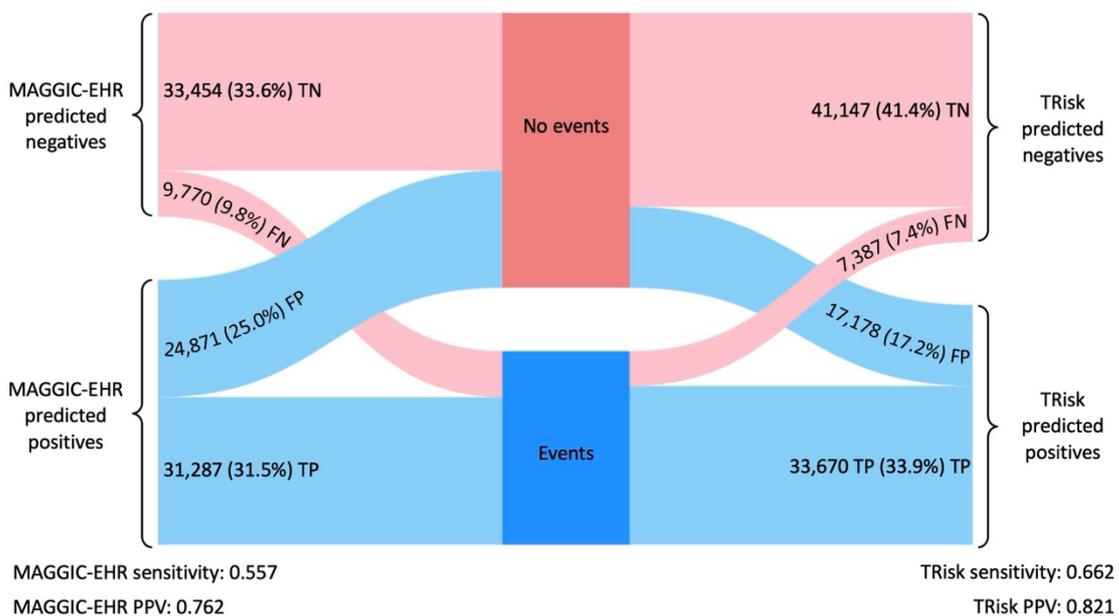

*Figure 3. **Impact analyses at the 50% decision threshold for (A) 12- and (B) 36-month all-cause mortality prediction on UK validation data.** Sankey diagrams compare predicted outcomes between the models, showing how patient classification compares to actual outcomes, "Events" (i.e., patients who suffered mortality) and "No events " (i.e., patients who did not suffer mortality) categories at 50% threshold (denoted as dark blue and red respectively). Light blue and red colours represent positive and negative predictions respectively. TP: true positive; TN: true negative; FP: false positive; FN: false negatives.*

Analyses of USA data

On the MIMIC-IV dataset (**Figure S6**), 21,767 patients were selected for analysis (13,060 patients in validation) with median follow-up of 7 (IQI: [2, 13]) months (**Table S7**). MIMIC-IV cohort had a similar age distribution, smoking patterns, and prevalence of diabetes, COPD, and atrial fibrillation to the CPRD cohort (**Table S7**).

In analyses of the validation dataset for 36-month mortality prediction, the transfer learning variant of TRisk demonstrated superior C-index of 0.802 (0.789, 0.816) compared to other TRisk modelling strategies with similar superiority in AUPRC (**Figure S7; Table S8**). Also, it was found to be well calibrated across the full spectrum of risk (**Figures S8; Table S9**) with appropriate stratification of high-risk individuals as evidenced by a peak in the high end of the risk spectrum (**Figure S8B**). Decision curve analyses demonstrated that the transfer learning TRisk variant provided greatest net benefit across all relevant thresholds (**Figure S8C**). All findings were preserved in analyses of 12-month mortality prediction (**Figure S9; Tables S8, S9;** extended results are in **Supplementary Results: Analyses on MIMIC-IV validation dataset**).

Explainability analyses

Our explainability analysis revealed that while contributions were modest, TRisk successfully identified validated risk factors as contributing to mortality risk (**Figure 4A**) across both cohorts. Among the top ten contributing encounters across cohorts (**Figure 4B**), two were consistently ranked highest in both cohorts (i.e., green lines), while the next five were identified as important in the USA validation with varying ranks (i.e., blue lines). Sex-stratified analyses aligned with the main findings, while age-stratified analyses showed encounter contributions generally increasing with age (**Figure S10**). In time-to-baseline analysis, most diseases showed higher contribution scores when closer to baseline, with diminishing association over time. Notably, all types of cancers maintained positive contributions even beyond 10 years before baseline, suggesting persistent mortality associations long after initial diagnosis across both cohorts.

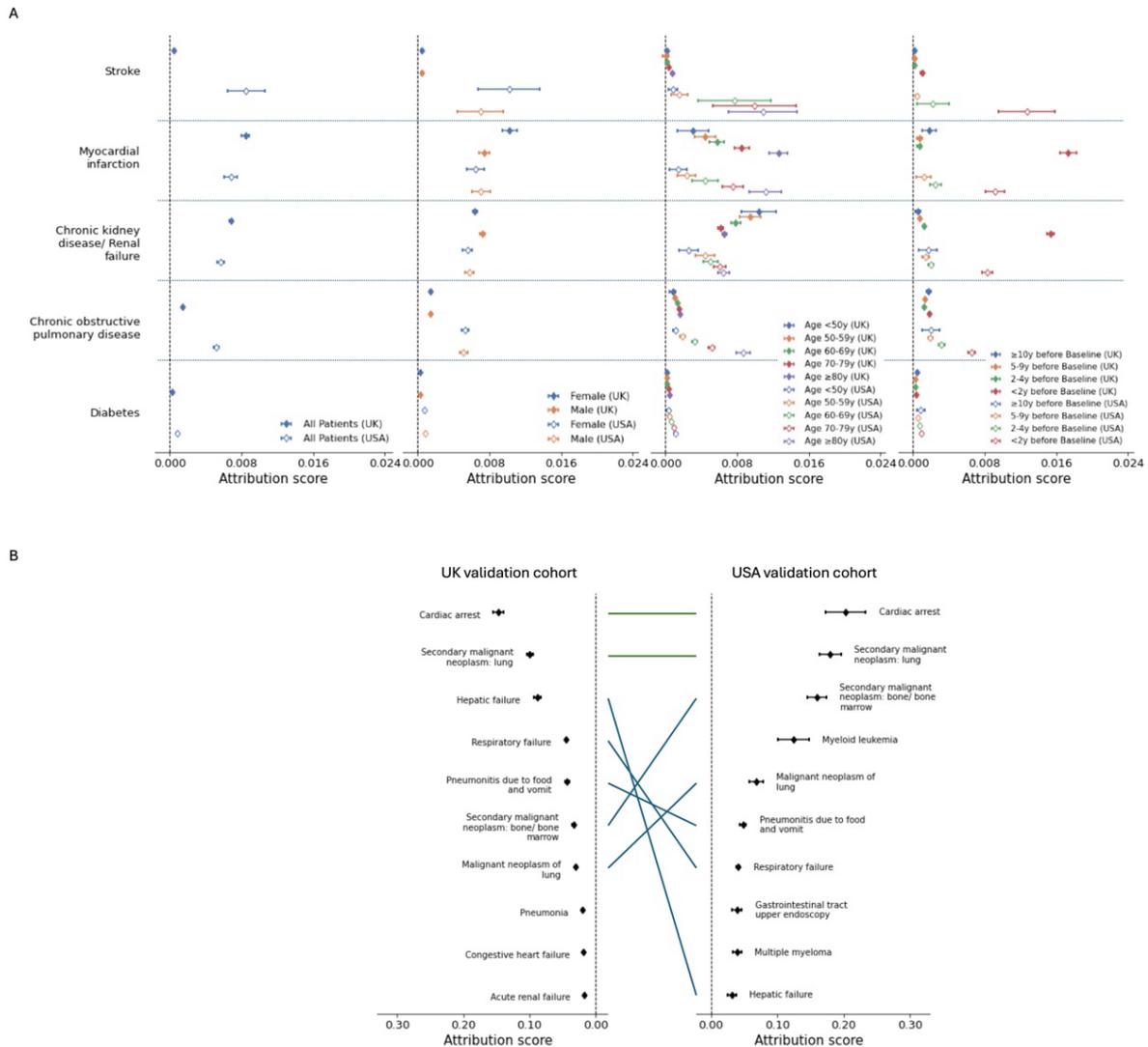

*Figure 4. Average contribution for (A) validated risk factors and (B) top ten encounters found as most contributing to mortality risk prediction on UK and USA validation cohort datasets. Point estimates of contribution values and associated 95% confidence intervals are presented for each encounter. For (A) average contribution for validated risk factors, average contribution scores are calculated with appropriate 95% confidence interval across all patients, stratified by sex, stratified by age at first encounter, and time before baseline of first encounter. For (B), we looked at encounters with top ten contribution scores as captured by our TRisk model across UK (left) and US (right) datasets. Green line denotes both encounter and rank were preserved between validation analyses; blue line denotes only encounter was preserved.*

### Analyses of secondary outcomes on UK data

On CPRD, we conducted analyses for predicting fatal and non-fatal cardiovascular outcomes (median follow-up of 5; IQI: [1, 16] months), cardiovascular-related mortality (9; IQI: [2, 28] months), and renal outcomes (6; IQI: [2, 20] months). Findings for prediction of 36- (**Figures S11-S14; Table S2, S3, S10-S12**) and 12-month (**Figures S15-S17; Table S5**) risk of these secondary outcomes were overall similar to findings for all-cause mortality analyses (details in **Supplementary Results: Analyses of all outcomes on CPRD validation dataset**).

### Discussion

As a model utilising solely routine EHR, TRisk demonstrated 10-20% higher C-index and 15-30% higher AUPRC than conventional models, with acceptable calibration across all outcomes at 12 and 36 months in the UK validation cohort. In impact analyses, the model reduced false positives and false negatives by up to 31% and 46% respectively compared to MAGGIC-EHR. Furthermore, TRisk transferred well from UK to USA EHR setting, with strong discrimination and calibration for mortality prediction. Finally, explainability analyses highlighted the model captured validated predictors, providing face validity to AI feature extraction processes. Moreover, it revealed some underappreciated risk factors such as cancers, hepatic failure, and respiratory conditions across both cohorts.

The rising global burden of HF demands refined management.[31,32] Current state-of-the-art risk models for HF cohorts have several limitations that have impeded their widespread clinical adoption. Univariable risk tools like NYHA class or LVEF have shown limited prognostic value.[33] Additionally, multivariable risk tools (**Table S13**), such as the Seattle HF Model, are often tailored for specific HF patient groups, like those with reduced EF.[5] Few generally applicable models exist (e.g., MAGGIC-EHR);[5,9] however, even they rely heavily on specialised imaging markers, indeed resource-intensive to procure. This creates barriers to implementation – particularly post-pandemic, where only 51% of hospitals meet the target of providing diagnostic echocardiography to 90% of HF patients.[34] Moreover, while these tests inform immediate care, their predictive value diminishes over time. This offers one explanation for why most models achieve modest discrimination (C-index <0.8) for medium-term outcome prediction with accompanying poor sensitivity/PPV.[8,13] Our implementation of MAGGIC amended for primary and secondary care EHR demonstrated similar findings. As a result, HF management guidelines have been hesitant to integrate such models into care pathways.

TRisk, a model that accounts for complex patient journeys, demonstrated superior discrimination across four outcomes in the UK validation study of HF patients whilst maintaining calibration. Importantly, we found no evidence that TRisk exhibits biased predictions across key subgroups, including different sexes, age ranges, and baseline characteristics. This unbiased approach could be seminal for preventing disparities in HF management, maintaining equity in patient care and mitigating the perpetuation of existing inequalities.

In our USA data study, we explored adapting complex AI models to a different setting. The smaller MIMIC-IV sample size proved insufficient for training TRisk "from scratch," but leveraging the larger CPRD dataset for initial training improved performance. Further fine-tuning on MIMIC-IV data via transfer learning enhanced discrimination (~0.04 C-index increase) and simultaneously aided in recalibration. With this generalisability study, we demonstrated successful transfer learning of TRisk across multiple dimensions: from (1) larger to smaller cohort, (2) primary/secondary to inpatient/outpatient setting, (3) UK to USA data.

Data-driven AI models like TRisk offer a novel approach to predicting HF outcomes by capturing the holistic patient health journey. Unlike conventional models requiring extensive resources to identify specific predictors, TRisk automatically discovers important patterns from routine EHR data.[14,16] The model excels with complex patient data and moreover, perhaps counter-intuitively is designed for cohorts with numerous clinical visits, comorbidities, and procedures - precisely where conventional statistical models reach their limitation.[9] This represents a significant advance in risk assessment for patients with complex diseases at baseline.

The explainability analyses further highlighted the importance of capturing complex patient journeys. While the model recognises established risk factors for mortality prediction, it identifies a broader range of conditions - particularly cancers, hepatic failure, and respiratory problems - as more significant predictors across both cohorts in line with recent research.[2,35,36] Additionally, temporal analysis of risk factors revealed that while recent clinical encounters generally carried greater predictive weight, cancer diagnoses maintained significant predictive value even ten years before baseline. This finding could reflect either direct effects of cancer or of cancer therapies - a distinction that warrants future research but aligns with current cardio-oncology literature.[35] These insights directly explain the drivers of TRisk's advantage over conventional models (e.g., MAGGIC), which omit longitudinal modelling of these and other identified predictors. Indeed, especially in complex patients with pre-existing diseases (e.g., HF, diabetes), these findings demonstrate the predictive value of capturing the full scope of a patient's health trajectory - from historical conditions like cancer to recent clinical events - rather than relying solely on traditionally understood markers.

TRisk's success promises significant clinical benefits throughout the HF care pathway without adding to the clinical workload by requiring additional information or specialised tests during consultations. At the population level, these risk scores can be used to audit HF service quality and assess the impact of service changes. At the individual level, TRisk can periodically estimate risk scores offline and flag patients at risk who may need further clinical attention, offering key advantages: better identification of low-risk patients could prioritise medication over transplantation, potentially reducing waitlists and healthcare costs,[37,38] while accurate detection of high-risk cases could enable timely palliative care referrals, reducing hospitalisations and improving quality of life.[39] Furthermore, TRisk has been validated for use across the spectrum of time-points following HF diagnosis. It can be utilised repeatedly, adapting to changes in patient health – whether from disease progression, treatment complications, or therapeutic interventions – by using updated EHR data to inform ongoing care decisions. Validated in both primary/secondary care (UK) and inpatient/outpatient care (USA) settings, TRisk can aid shared decision-making for HF management.

Strengths and limitations.
We comprehensively validated TRisk across two datasets, assessing performance at 12 and 36 months as per guideline recommendations.[13] We analysed the model's decision-making processes by examining influential medical encounters and discovered previously overlooked mortality risk factors. Finally, we demonstrated its versatility by testing on three additional outcomes using UK health records.

In terms of limitations, the MAGGIC model was not designed for use on routine EHR data. Specifically, LVEF measurements were fully absent while NYHA classifications variable suffered from missingness. However, given the scarcity of risk prediction models for HF patients in routine EHR, we developed MAGGIC-EHR as a benchmark for comparison with TRisk. We also implemented MAGGIC-EHR+, which includes additional important predictors, to ensure fair comparison with AI approaches. Moreover, there is no evidence to suggest that our MAGGIC-EHR models prognosticates materially differently than the originally model; in fact, predictive performance of our implementation aligns with previous MAGGIC validation studies (C-index: ~0.73; **Table S13**).

Conclusion
TRisk achieved superior prediction of clinical outcomes in HF patients using only routine EHR data, outperforming conventional models without sacrificing calibration. The model's robust performance translated successfully to USA healthcare settings through transfer learning, demonstrating its potential for broader application. The explainability analyses revealed how risk profiles evolve over time, with both recent factors like hepatic failure and long-term influences like cancer driving prognostication. Utilising TRisk as an offline prognostication tool that periodically estimates risk scores could improve audit of current and future HF services and refine care decisions at the individual level.

# Supplementary Material

A Transformer-based survival model for prediction of all-cause mortality in heart failure patients: a multi-cohort study

Shishir Rao, DPhil[1], Nouman Ahmed, MSc[1], Gholamreza Salimi-Khorshidi, DPhil[1], Christopher Yau, DPhil[1,2], Huimin Su, MPhil[3], Nathalie Conrad, DPhil[1,3], Folkert W Asselbergs, MD PhD[2,4,5], Mark Woodward, PhD[6,7], Rod Jackson, PhD[8], John GF Cleland, MD[9], Kazem Rahimi, DM[1*]

[1] Nuffield Department of Women's & Reproductive Health, University of Oxford, Oxford, United Kingdom
[2] Health Data Research UK, London, UK
[3] Department of Cardiovascular Sciences, Katholieke Universiteit Leuven, Belgium
[4] Amsterdam University Medical Center, Department of Cardiology, University of Amsterdam, Amsterdam, The Netherlands.
[5] Institute of Health Informatics, University College London, London, UK
[6] The George Institute for Global Health, University of New South Wales, Newtown, NSW, Australia
[7] The George Institute for Global Health, Imperial College London, London, United Kingdom
[8] School of Population Health, Faculty of Medical and Health Sciences, University of Auckland, Auckland, New Zealand
[9] British Heart Foundation Centre of Research Excellence, School of Cardiovascular and Metabolic Health, University of Glasgow, Glasgow, UK

## Table of Contents

# Supplementary Methods

## Supplementary Methods: Clarification on CPRD validation study

In this study, patients from different general practices (GPs) formed the derivation and validation datasets. Derivation data was used for model fitting/training while the validation data was used for validation of the models. For TRisk, to avoid overfitting in the training process, which is a concern with high-dimensional and AI parametric modelling, 5% of the derivation dataset was randomly selected to conduct end-of-epoch testing. This testing at the end of each epoch (i.e., one full iteration of training in which the model has trained on all patient selected for training) was used to ensure the model has not overfit on training data (i.e., ultimately aiming to avoid data "memorisation"). After AI model training on the 95% of patients in the derivation dataset was estimated to have converged, the AI model was evaluated on the validation dataset. All patients in derivation dataset were used for the fitting of the conventional statistical modelling solutions.

## EHR pre-processing and AI modelling

For our AI modelling, we converted all diagnoses recorded by GPs from the Medcode diagnostic codes to the level 4 ICD-10 codes (i.e., 1 number following decimal point like N29.9, J12.2, I50.1), ensuring uniformity in coding across hospital and general practice records. This conversion utilised a phenotyping dictionary from NHS Digital and SNOMED-CT.[1] The medications were recorded in the format of the native CPRD product code format. In which case a British National Formulary (BNF) mapping was applicable, we translated the relevant product codes to their corresponding BNF codes, retaining only section-level codes (i.e., first four digits of the BNF coding format).[2] For product codes that were not able to be mapped with the BNF's mapping algorithm, we used NHS Digital maps which mapped product codes to the appropriate virtual therapeutic moiety (VTM) codes, a vocabulary set for abstract representations of medication ingredients maintained and updated by the SNOMED-CT group.[1] Consequently, our final medication code list included both BNF and VTM codes. In contrast, procedure codes, recorded as Office of Population Censuses and Surveys (OPCS) Classification of Interventions and Procedures version 4 codes, were maintained in their original format without undergoing any additional mapping or transformation. All codes that were prevalent in at least 0.1% of CPRD Aurum were used for modelling.

To understand the evolution of BEHRT into the TRisk model (**Figure S1**), an understanding of the BEHRT model is essential. BEHRT comprises of three primary elements: (1) An embedding layer that incorporates three layers of raw EHR data for modelling (see below for more details), (2) a Transformer-based feature extractor, which extracts rich features from temporal EHR data utilising the multi-head self-attention mechanism, and (3) a binary prediction layer.[3] This layer uses a sigmoid activation function followed by a linear transformation of the latent patient representation vector to produce risk assessments (i.e., the likelihood of an outcome occurring) for an individual patient.

Regarding the (1) embeddings, the input embedding layers encompass encounter data, age, and positional encoding. The encounter layer includes diagnostic, medication, and any other datapoints from raw EHR (e.g., procedures), representing a patient's medical history up to the baseline. The age layer specifies the exact age at the time of each encounter, calculated as a simple subtraction between the event date and estimated birth date; for the purposes of pseudonymisation, the Clinical Practice Research Datalink (CPRD) supplies only the year of birth; therefore, the birth date is assumed to be July 1st of the stated birth year.[4] Positional encoding consists of pre-set embeddings that impart sequential context to the Transformer model. Pivotally, these positional embeddings denote visit number for a particular encounter; the encounters captured in first visit will be given position encoding corresponding to visit 1, those for second visit, the encoding to visit 2, and so on. The combined embeddings from all three features form the comprehensive high-dimensional vector representation of each encounter.

The TRisk model makes several changes on the underlying, BEHRT model to transform it into an accurate survival model. First, the age embedding now explicitly includes age at baseline to more expressively capture time of prediction. In this way, for a particular patient, prediction at age, 61 years, will be different than at age, 65 years, and hence will be appropriately denoted differently. Also, in this way, baseline age is explicitly included as a variable for use in prediction of the outcome of interest.

Second, the embedding layers are not summed like in previous iterations of BEHRT. Rather, the three separate embedding layers are concatenated using tensor concatenation operations and multiplied by a weight vector that transforms the space: 3•E → E (i.e., E being the space of one embedding layer; 3E is three stacked layers). Finally, a non-linear, hyperbolic tangent functional transformation is applied to the product to enable more expressive latent representation of input raw EHR and ultimately, mitigate overfitting.

Third, TRisk utilises the Scalable Continuous-Time Survival Model through Ordinary Differential Equation Networks (SODEN) framework for modelling. Instead of maximum likelihood estimation (MLE) frameworks that utilise costly integral calculations for model training on censored data, the SODEN framework alternatively poses MLE as a differential-equation constrained optimisation objective.[5] Furthermore, unlike previous proportional hazard frameworks that have not been shown to be theoretically robust for stochastic gradient descent (SGD) using mini-batching (i.e., random non-overlapping subsets of the derivation dataset), the SODEN framework alleviates many issues of established survival model frameworks that have trouble scaling to "big data". Specifically, with ordinary differential equations modelling the time-to-event distribution, the framework for MLE on censored data becomes more flexible (i.e., absent of strong structural assumptions of the shape of the survival/hazard distribution), more appropriately scalable (i.e., can be trained using SGD).

Fourth and finally, the TRisk objective or cost function utilises an Explicit Calibration (EC) method for optimising for D-calibration, an omnibus measure of calibration, in addition to empirical loss.[6] With these updates, the TRisk model is theoretically more robust for scalable training on large-scale EHR data and ensures greater expressiveness and flexibility to ultimately ensuring optimal discrimination and calibration.

The TRisk model similar to the BEHRT model utilises the prediction token ('Predict token') to serve as a dense patient representation vector for each patient. This vector functioning as a dense feature vector is then fed into the SODEN framework survival modelling network that conducts risk prediction for a particular outcome.

### *Implementation of MAGGIC-EHR models*

The MAGGIC-EHR class of Cox models has been tailored for use on primary and secondary care administrative EHR in this work.[7] The raw variables extracted for MAGGIC-EHR modelling included sex, age, smoking status, diabetes, systolic blood pressure, creatinine, body mass index, New York Heart Association (NYHA) Functional Class, chronic obstructive pulmonary disease, any use of beta-blockers, any use of ACE inhibitor/ Angiotensin receptor blockers, incident HF diagnosis ≥18 months prior to baseline, and HF subtype (i.e., preserved, reduced, or unknown according to previously published method[8]). In addition, the age and systolic blood pressure interaction terms for the HF subtype were included into the model. For the extended MAGGIC-EHR model with additional predictors, referred to as MAGGIC-EHR+ model, sodium level measurement, atrial fibrillation, stroke, myocardial infarction, history of percutaneous coronary intervention, history of coronary artery bypass grafting procedure were extracted additionally. For both benchmark models, the predictors were captured from both primary care and HES records. Average of records for systolic blood pressure, creatinine, body mass index, and sodium that were collected in the 36 months leading up to baseline were used as baseline variables for modelling. Last known smoking status in the 36 months leading up to baseline was used as baseline smoking status.

Missing values for systolic blood pressure, smoking status, body mass index, creatinine, and NYHA Functional Class were imputed for the MAGGIC-EHR model. Separate imputation was conducted for the MAGGIC-EHR+ model accounting for missing sodium measurements. Procedurally, for imputation of all-cause mortality prediction MAGGIC-EHR and MAGGIC-EHR+ models, all variables for each model were included for imputation along with Nelson-Aalen estimate of baseline cumulative hazards for the outcome of mortality. Imputations were conducted on each derivation and validation datasets of our cohort separately using the "mice: Multivariate Imputation by Chained Equations" package in R.[9] Following the extraction of derivation and validation datasets, the statistical models were fit on each of the five imputed derivation datasets with appropriate predictors (i.e., 5 MAGGIC-EHR/MAGGIC-EHR+ models) for prediction of the primary outcome, all-cause mortality. The five Cox models were pooled using Rubin's rules and evaluated on the 5 imputed validation datasets. The predictions were also appropriately pooled using extensions of the Rubin's rules (i.e., complementary log-log transformations) and utilised for downstream analysis. The same procedures for model fitting and evaluation on validation datasets were repeated for the analyses of other outcomes: (1) fatal and non-fatal cardiovascular event prediction, (2) cardiovascular-related mortality prediction, and (3) renal outcomes (i.e., chronic kidney and end-stage renal diseases) prediction.

It must be noted that the original MAGGIC model derivation study has solely investigated 12- and 36-month all-cause mortality.[7] To serve as the benchmark conventional model in this work, we used MAGGIC-EHR to not only predict 12- and 36-month all-cause mortality but also to demonstrate benchmark performance for the secondary outcomes investigated. Interestingly, while MAGGIC was not derived for secondary outcome investigations, validation studies have indeed been conducted for other outcomes such as CV-related mortality (more details in **Table S13**).[10]

### Implementation of AI models

For the TRisk class of AI models, the implementation was carried out in Python programming language using the graphical processing unit (GPU) compatible PyTorch coding framework. One NVIDIA A100 Tensor Core GPU was used for each of the model training, validation, and transfer learning pipelines. For optimiser, the Adam optimiser was used as conventional with the BEHRT class of models; learning rates were optimised during the course of model training using the exponential decay method.[3]

### Supplementary Methods: Clarification on MIMIC-IV validation study

In additional external validation analysis on the MIMIC-IV dataset, we first identified cohort of HF patients using established ICD-10 codes for HF identification.[11] We identified patients with HF between the ages of 40 and 90 in the years between 2008 and 2019. Similar to cohort selection in CPRD, the index date (baseline) was randomly selected from the eligible patient time period for each individual and earliest of 12-month after last hospital visit (i.e., this is date after which patient is lost-to-follow-up) and death. Patients were censored at the earliest of the 12-month mark after last hospital admission visit, outcome of interest (i.e., death), 4 years after the index date (i.e., truncation after 48-month mark), or the study end date (1 January 2019). Mortality data were provided by linkage to Massachusetts, USA state mortality records.

On a random selection 60% of the cohort, we conducted external validation of the TRisk model, henceforth referred to as the MIMIC-IV external validation dataset. We used the remaining 40% of the identified cohort for transfer learning and fine-tuning experiments, henceforth referred to as the MIMIC-IV fine-tuning dataset. Within the 40% cut of the dataset, for experiments involving TRisk model fitting/training, 10% of the dataset was randomly selected to conduct end-of-epoch testing. This testing at the end of each epoch (i.e., one full iteration of training in which the model has trained on all patient selected for training) was used to ensure the model has not overfit on training data (i.e., ultimately aiming to avoid data "memorisation"). For statistical model fitting, all patients in the MIMIC-IV fine-tuning dataset were used. All models were evaluated on the MIMIC-IV external validation dataset.

### Supplementary Methods: Clarification on modelling on MIMIC-IV dataset

We evaluated three different models on the validation dataset of MIMIC-IV[11] (see above for more details): TRisk trained on CPRD, TRisk trained "from scratch" on fine-tuning dataset, and TRisk trained on CPRD, and transferred and further fine-tuned on the fine-tuning dataset.

As TRisk requires diagnosis, medication, and procedure records to be encoded in accepted vocabularies (e.g., ICD-10 for diagnosis records), we conducted the mapping for MIMIC-IV dataset. Specifically, since the TRisk needed to be successfully transferred from CPRD to MIMIC-IV, raw records from MIMIC-IV needed to be mapped to the same vocabulary sets: (1) ICD-10 for diagnoses, (2) mixed, BNF and VTM for medications, and (3) OPCS for procedures. For diagnoses records, the MIMIC-IV dataset provides records in a mixture of ICD-9 and ICD-10 codes. ICD-9 codes were mapped to ICD-10 codes using established maps provided by the public code repository, "MIMIC-IV-Data-Pipeline", the MIMIC-IV data processing pipeline developed by Gupta et al.[12] For medication records, we used SNOMED CT maps provided by Athena, the Observational Health Data Sciences and Informatics (OHDSI) vocabularies repository to map the National Drug Code (i.e., drug encoding vocabulary in US) to BNF and VTM DM+D (i.e., SNOMED) codes.[13] For the procedures, encoded in ICD-9 and ICD-10 Procedure Coding System (PCS), we mapped them to SNOMED universal procedure codes and finally OPCS using Athena vocabulary mapping libraries.[13]

*Supplementary Methods: Clarification on explainability analyses*

In this work, we used integrated gradients to understand predictive process of the TRisk model.[14] In gist, the integrated gradient method is a method that helps explain how AI models make decisions by tracking the importance of each input feature along a straight path from a baseline to the actual input. Starting from a neutral reference point – a zero-value vector for input to our TRisk model – it measures how much each feature contributes to the model's final prediction by gradually transforming this baseline into the actual input. By averaging these contributions along the path, the integrated gradients method provides a clear measure of which parts of the input were most influential in the model's decision, making complex AI models like TRisk more interpretable to humans.

In our work, we adapted the integrated gradients method to work within the SODEN framework, calculating how each patient encounter influences the model's output. For repeated encounters (e.g., multiple diagnoses of the same condition or multiple prescriptions of same medication), we used the highest contribution value across all instances. We focused our analysis on encounters that appeared in at least 1% of patients to ensure statistical reliability. In age-based analyses, for each encounter, we grouped patients into subgroups based on age of first recording of the particular encounter. Similarly, for the analyses of time between recording of encounter and baseline, we grouped patients into subgroups based on first (as opposed to any other subsequent recording) recording of encounter and baseline.

## Supplementary Results

### Supplementary Results: Analyses of all outcomes on CPRD validation dataset

For all-cause mortality prediction, TRisk demonstrated significantly higher C-index and AUPRC as compared to MAGGIC-EHR class of models (**Figure S3; Table S2**). Statistical models were similar across both metrics (**Figure S3; Table S2**). In terms of predictive distribution, the AI approach clearly demonstrated two peaks in the lowest and highest ends of the spectrum of predicted risk implying more nuanced stratification of risk as compared to the MAGGIC-EHR class of models (**Figures 2B, S4B**). Decision curve analysis showed that TRisk provided significantly greater net benefit than both MAGGIC-EHR and MAGGIC-EHR+ models (**Figure S4C**) across the spectrum of clinically relevant thresholds (0.0-0.6). The subgroup discrimination analyses demonstrated that the deviation from overall cohort discrimination was mitigated for TRisk as compared to benchmark statistical modelling solutions (**Table S3**). 12-month model performance captured similar trends found in 36-month analyses (**Figures S5; Table S5**).

In analyses of secondary outcomes, overall and subgroup discrimination, calibration, and decision curve trends for models predicting both 36- (**Figures S11-S14; Table S2, S3, S10-S12**) and 12-month (**Figures S15-S17; Table S5**) risk were generally similar to trends captured in analysis of the main outcome.

Significant advancements in AI modelling were key for securing TRisk's gains in discrimination whilst remaining well-calibrated. TRisk with the SODEN framework and the enhanced EHR data representation layer outperformed conventional models. The SODEN framework offered more flexible modelling of the outcome space as compared to the proportional hazards framework (i.e., the framework used in Cox models and DeepSurv models), crucial for complex cohorts such as those with HF.[5] Additionally, TRisk's nuanced non-linear embedding module enabled extraction of more informative and richer features from minimally processed EHR data than predecessor models. In terms of calibration, TRisk demonstrated acceptable calibration, as evidenced by visual and ICI metrics. In summary, the fusion of the (1) revised EHR representation layer, (2) Transformer-based SODEN framework, and (3) explicit calibration regularisation scheme for improved calibration underscores the novelty and effectiveness of TRisk.

### Supplementary Results: Impact analyses on CPRD validation dataset

In impact analysis (**Figure 3; Table S6**), the benchmark MAGGIC-EHR model predicted ≥ 50% mortality rate in the 12 months following baseline for 13,042 of the 31,634 who actually died (i.e., sensitivity: 0.412) with 9,870 false alarms (i.e., PPV: 0.569). On the other hand, TRisk captured 1.7 times more correct cases with 21,586 accurate predicted cases (i.e., recall: 0.682) and predicted 699 fewer false alarms (i.e., PPV: 0.702).

Similarly, evaluating models with ≥ 50% predicted mortality rate in the 36 months following baseline, TRisk outperformed the benchmark approach with 0.105 and 0.059 greater PPV and sensitivity respectively. Importantly, TRisk demonstrated improvements in PPV and sensitivity metrics all whilst identifying 5,310 fewer at-risk patients than MAGGIC-EHR. Ultimately, TRisk enabled more accurate prediction with both fewer false positives and negatives as compared to benchmark MAGGIC-EHR approach.

### Supplementary Results: Analyses on MIMIC-IV validation dataset

In analyses of the validation dataset for 36-month mortality prediction, the transfer learning variant of TRisk demonstrated superior C-index of 0.802 (0.789, 0.816) compared to other TRisk modelling strategies with similar superiority in AUPRC (**Figure S9; Table S8**). TRisk without prior trained weights (i.e., randomly initialised), solely trained on MIMIC-IV fine-tuning data performed akin to a random, "coin-flip" risk model (i.e., ~0.5 C-index). In terms of calibration, while externally validated TRisk suffered from underestimation of risk at thresholds upwards of 0.4, the transfer learned TRisk was found to be better calibrated across the full spectrum of risk (**Figures 4, S8; Table S9**) with better stratification of high-risk individuals as evidenced by a peak in the high end of the risk spectrum (**Figure 4B**). Decision curve analyses demonstrated that the transfer learning TRisk variant provided greatest net benefit across all relevant thresholds (**Figure S8**). All findings were preserved in analyses of 12-month mortality prediction (**Figures S9; Tables S8, S9**).

Supplementary Figures

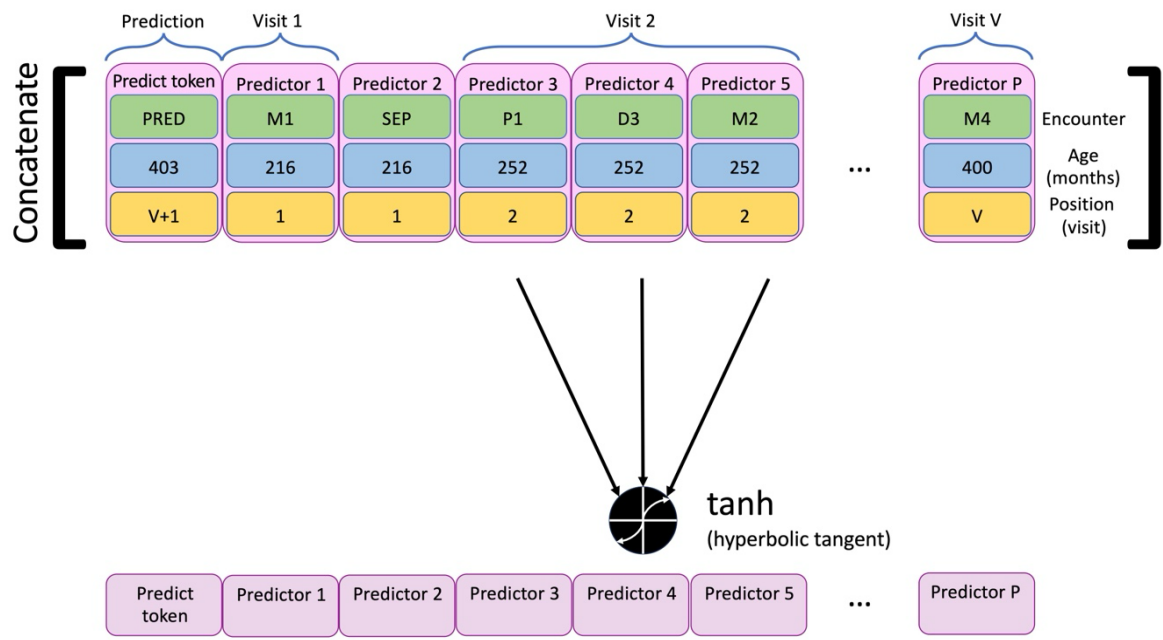

*Figure S1. TRisk input space for a hypothetical patient's medical history.*
For a hypothetical patient, the figure presents a sequence of medical history records represented by a group of three variables. Each record is comprised of an encounter (e.g., "M1" representing a hypothetical medication), the age at which the encounter was recorded in years (e.g., "216" months), and the visit number (e.g., visit "1"). These three components of the raw electronic health record data are represented by embeddings inputted into the model. The first character of the encounter code, D, M, and P represent diagnosis, medication, and procedure records respectively represented by a code in this figure. The number following the letter illustrates a hypothetical code of that modality type (e.g., 'D3' might represent code "I51.2" in ICD-10 encoding). "SEP" represents a separation character given to the model to split up the data between visits. Lastly, the "PRED" token (i.e., "predict token") is a special token that is used for prediction of the outcome (e.g., all-cause mortality outcome prediction). This special token has the age at baseline (e.g., "403" months at baseline for this hypothetical patient). As this is the visit that will happen following the final visit (i.e., visit number "V"), the visit for the predict token is appropriately enumerated as visit number "V+1". This predict token is inputted as displayed and following further transformation by Transformer architecture layers, the corresponding output state token will be used as a condensed latent patient representation layer for input into the ordinary differential equation-based survival prediction network layers.

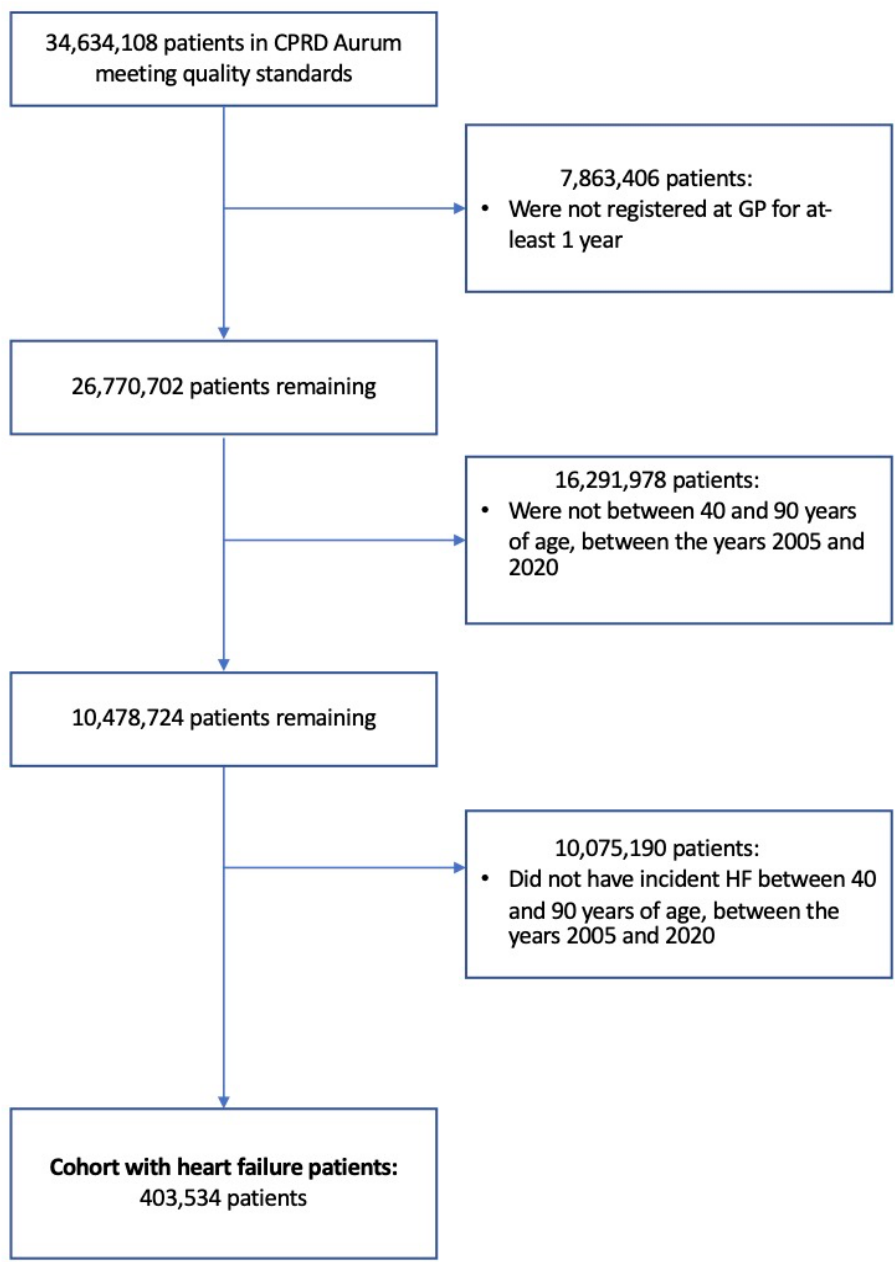

*Figure S2. Cohort selection flowchart for Clinical Practice Research Datalink (CPRD) Aurum dataset.*

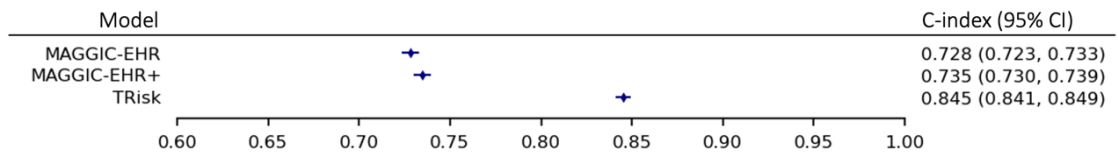

*Figure S3. Discriminative performance of models for 36-month risk prediction of all-cause mortality on UK validation data.*
*Discrimination is provided in this forest plot as assessed by C-index with 95% confidence intervals (CI) for various outcome investigations at 36-month timepoint.*

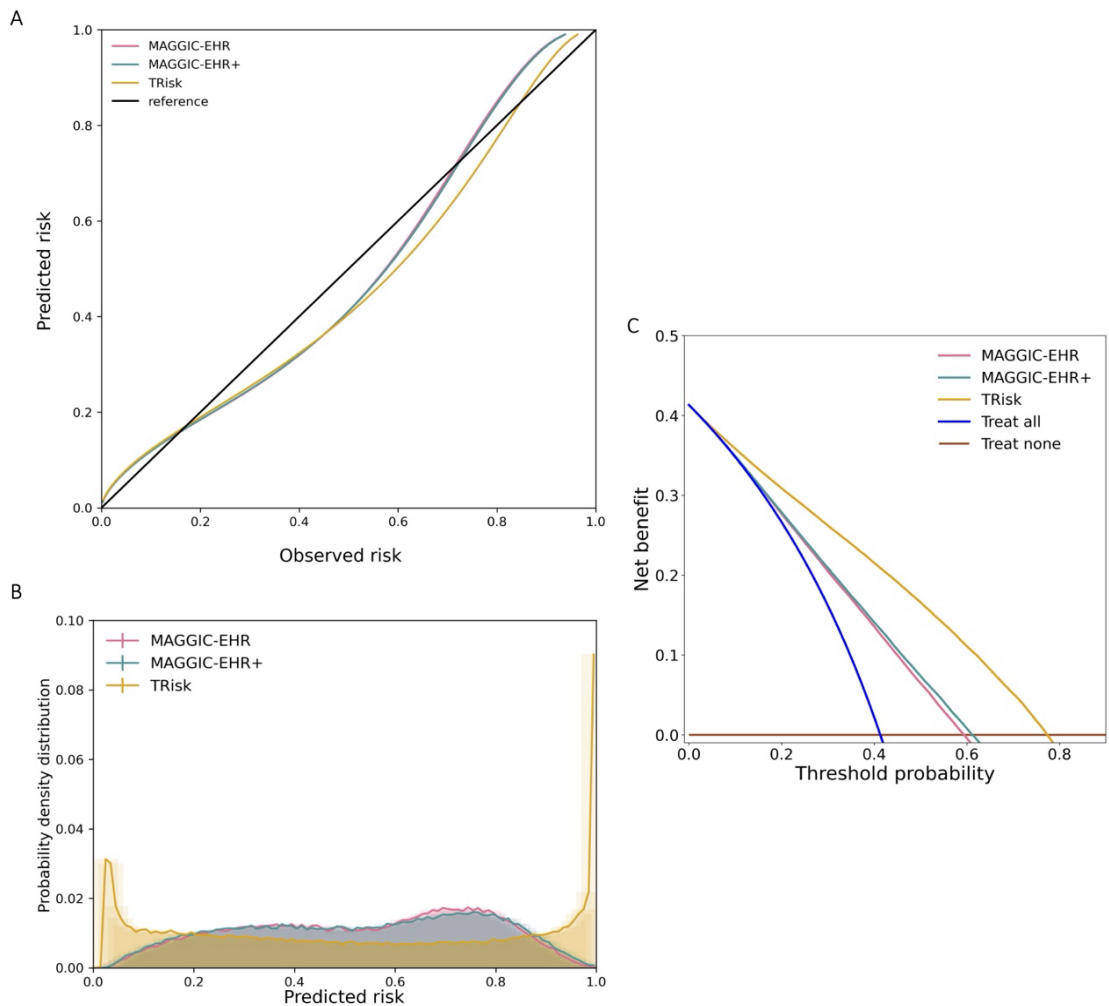

*Figure S4. Calibration curves, distribution of predicted risk of models, decision curve analyses for 36-month all-cause mortality across all models on UK validation data.*
*(A) Calibration curves, (B) distribution of predicted risk, and (C) decision curve analyses are presented for all models. Decision curve analysis (including censored observations) has been conducted for all models. Threshold probability is shown on the x-axis and the net benefit, a function of threshold probability, is shown on the y-axis and is the difference between the proportion of true positives and false positives weighted by odds of the respective decision threshold.*

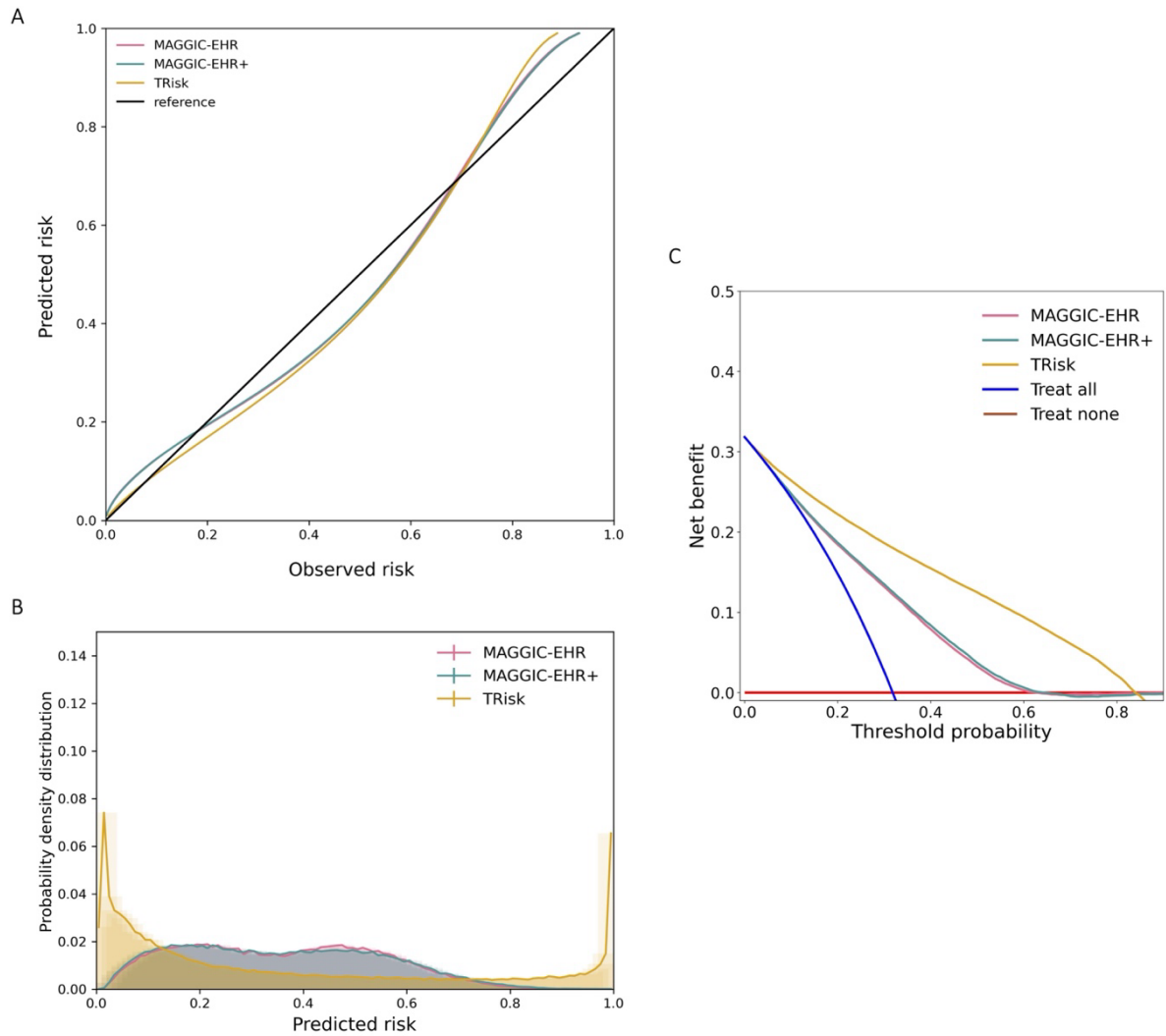

*Figure S5. Calibration curves, distribution of predicted risk of models, decision curve analyses for 12-month all-cause mortality across all models on UK validation data.*
*(A) Calibration curves, (B) distribution of predicted risk, and (C) decision curve analyses are presented for all models. Decision curve analysis (including censored observations) has been conducted for all models. Threshold probability is shown on the x-axis and the net benefit, a function of threshold probability, is shown on the y-axis and is the difference between the proportion of true positives and false positives weighted by odds of the respective decision threshold.*

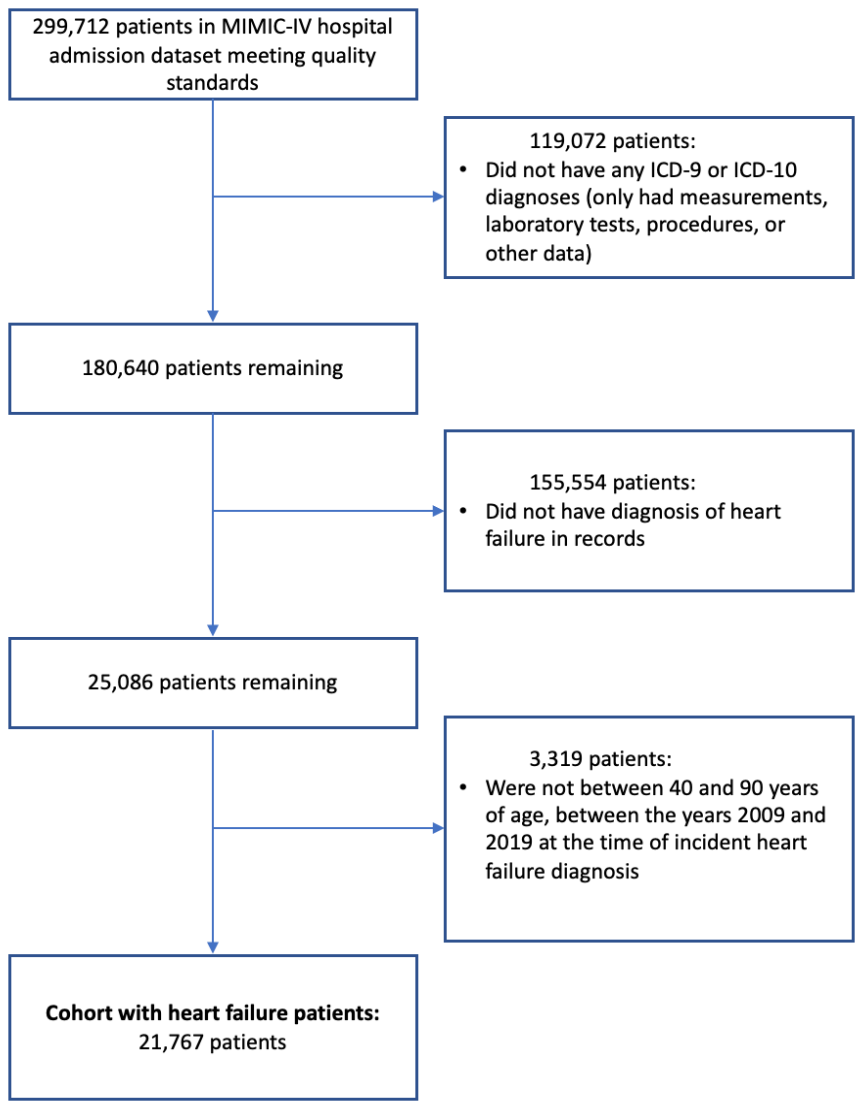

*Figure S6. Cohort selection for MIMIC-IV hospital admissions validation dataset.*
*MIMIC-IV: Medical Information Mart for Intensive Care-IV dataset.*

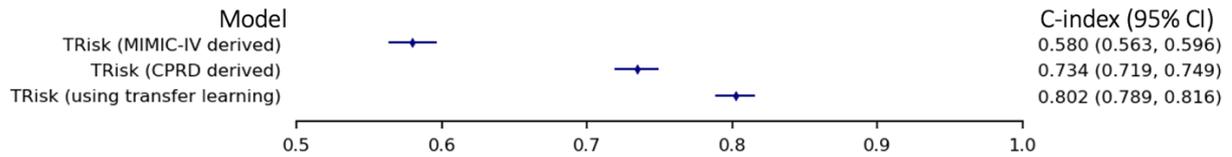

*Figure S7. Discriminative performance of models for 36-month risk prediction of all-cause mortality on USA validation data.*

*Discrimination is provided in this forest plot as assessed by C-index with 95% confidence intervals (CI) for various outcome investigations at 12-month timepoint. "TRisk (MIMIC-IV derived)" is randomly initialised and trained on MIMIC-IV fine-tuning dataset. "TRisk (CPRD derived)" is trained on CPRD derivation cohort. "TRisk (using transfer learning)" is trained on CPRD derivation cohort and fine-tuned on MIMIC-IV fine-tuning dataset. All models are validated on MIMIC-IV validation dataset.*

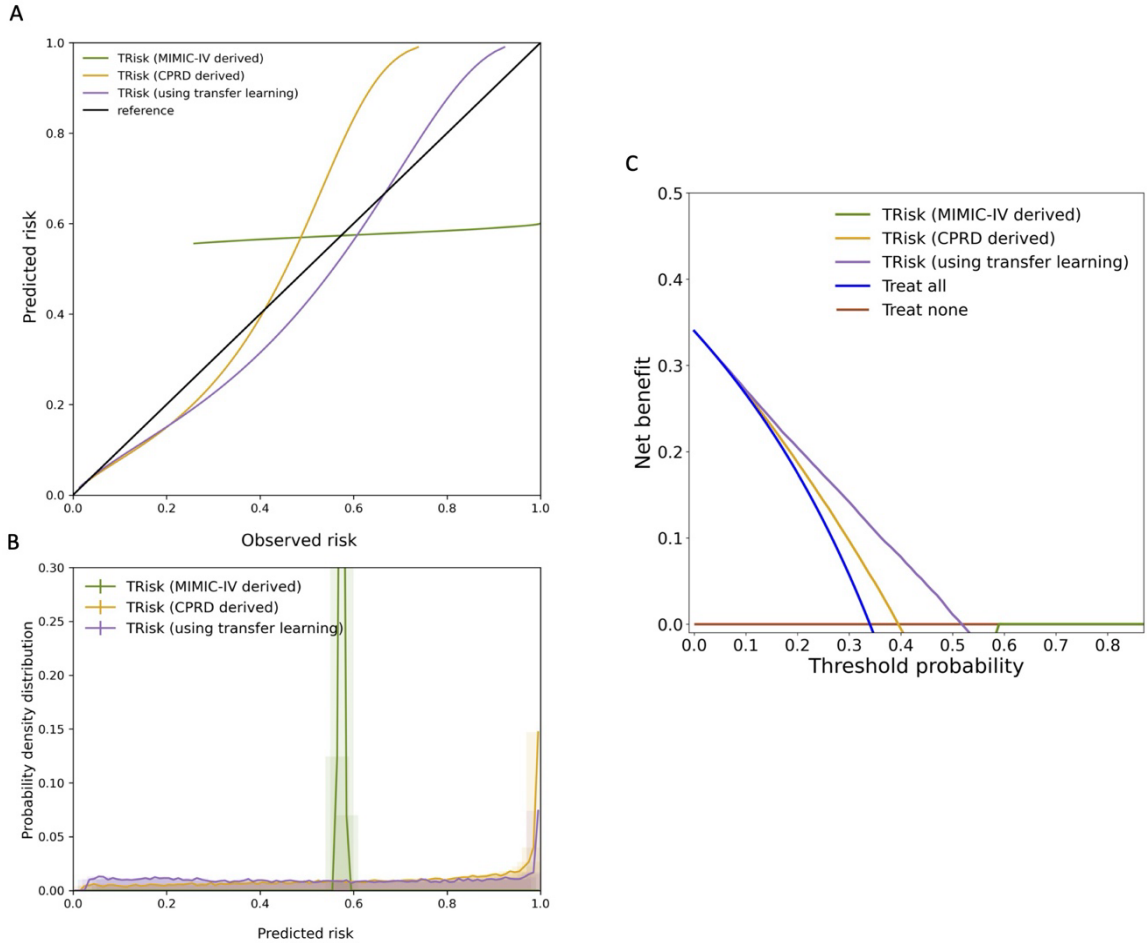

*Figure S8. Calibration curves, distribution of predicted risk of models, and decision curve analysis for 36-month all-cause mortality risk prediction on USA validation data.*
*(A) Calibration curves, (B) distribution of predicted risk, and (C) decision curve analyses are presented for all models. Decision curve analysis (including censored observations) has been conducted for all models. Threshold probability is shown on the x-axis and the net benefit, a function of threshold probability, is shown on the y-axis and is the difference between the proportion of true positives and false positives weighted by odds of the respective decision threshold. TRisk (MIMIC-IV derived) is randomly initialised and trained on MIMIC-IV fine-tuning dataset. TRisk (CPRD derived) is trained on CPRD derivation cohort. TRisk (using transfer learning) is trained on CPRD derivation cohort and fine-tuned on MIMIC-IV fine-tuning dataset. All models are validated on MIMIC-IV validation dataset.*

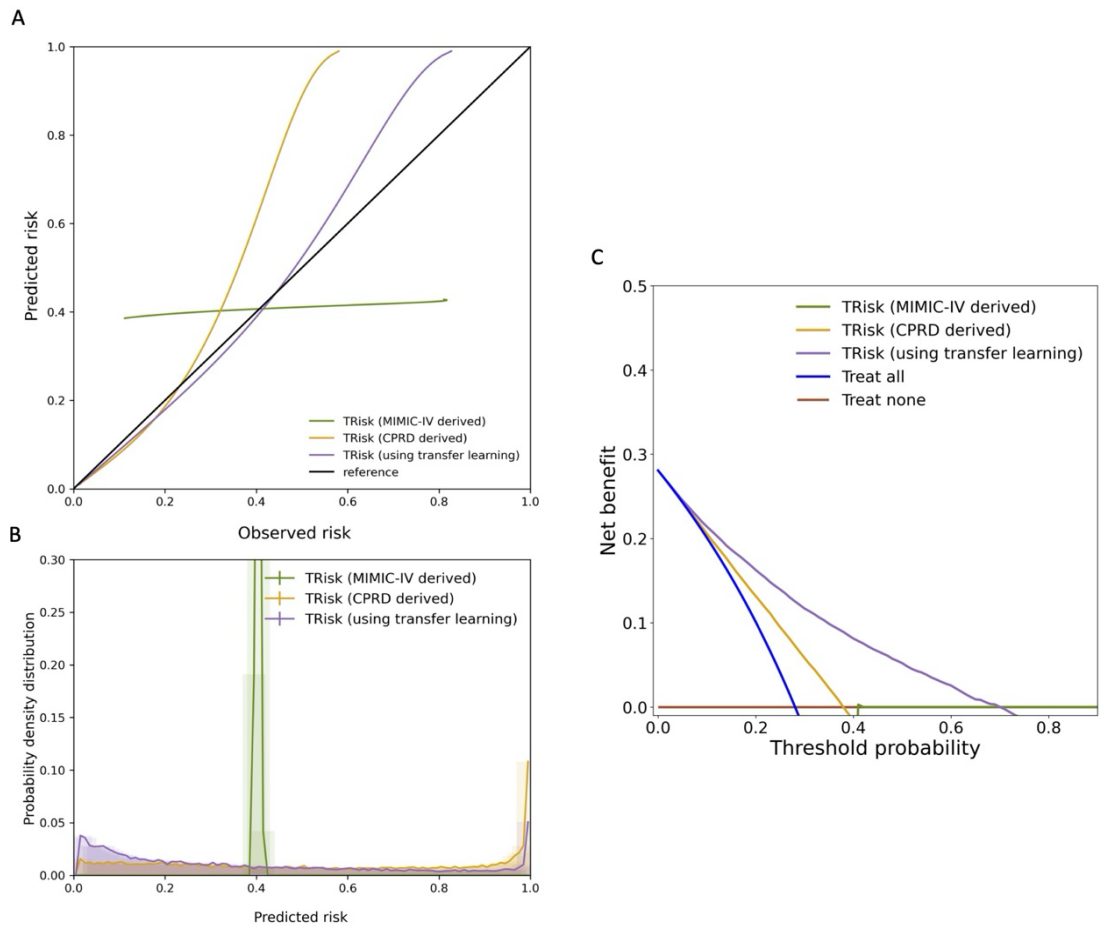

*Figure S9. Calibration curves, distribution of predicted risk of models, and decision curve analysis for 12-month all-cause mortality risk prediction on USA validation data.*
*(A) Calibration curves, (B) distribution of predicted risk, and (C) decision curve analyses are presented for all models. Decision curve analysis (including censored observations) has been conducted for all models. Threshold probability is shown on the x-axis and the net benefit, a function of threshold probability, is shown on the y-axis and is the difference between the proportion of true positives and false positives weighted by odds of the respective decision threshold. TRisk (MIMIC-IV derived) is randomly initialised and trained on MIMIC-IV fine-tuning dataset. TRisk (CPRD derived) is trained on CPRD derivation cohort. TRisk (using transfer learning) is trained on CPRD derivation cohort and fine-tuned on MIMIC-IV fine-tuning dataset. All models are validated on MIMIC-IV validation dataset.*

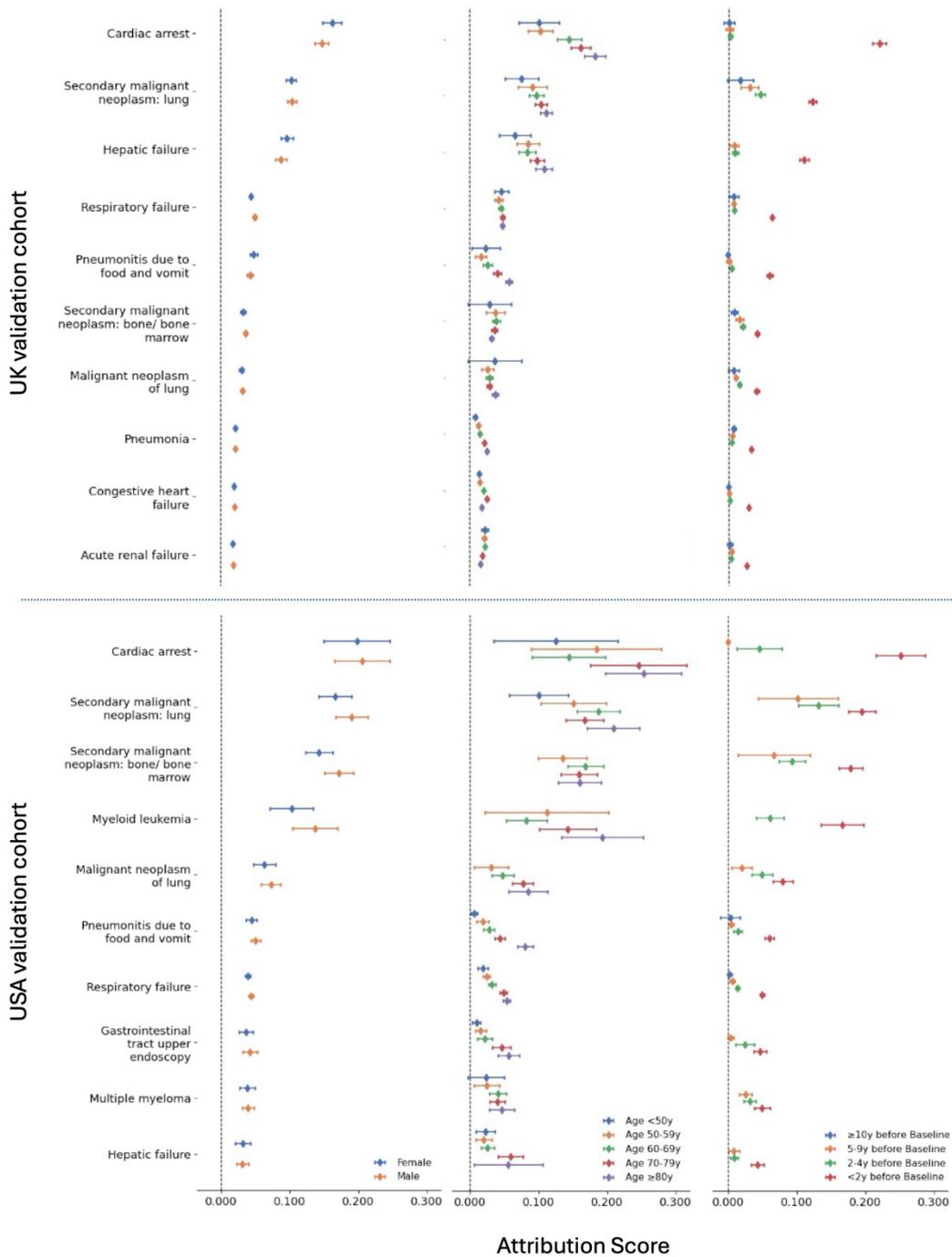

*Figure S10. Stratified analyses of top ten contributing encounters for mortality risk prediction on UK and USA validation datasets*

*For all top ten contributing encounters, we investigated contributions of top 10 contributing encounters in patient subgroups. Specifically, we investigated contribution patterns in both UK (top) and USA (bottom) validation cohorts stratified by sex, stratified by age at first encounter, and time between first recording of encounter and baseline. Point estimates of contribution values and associated 95% confidence intervals are presented for each encounter.*

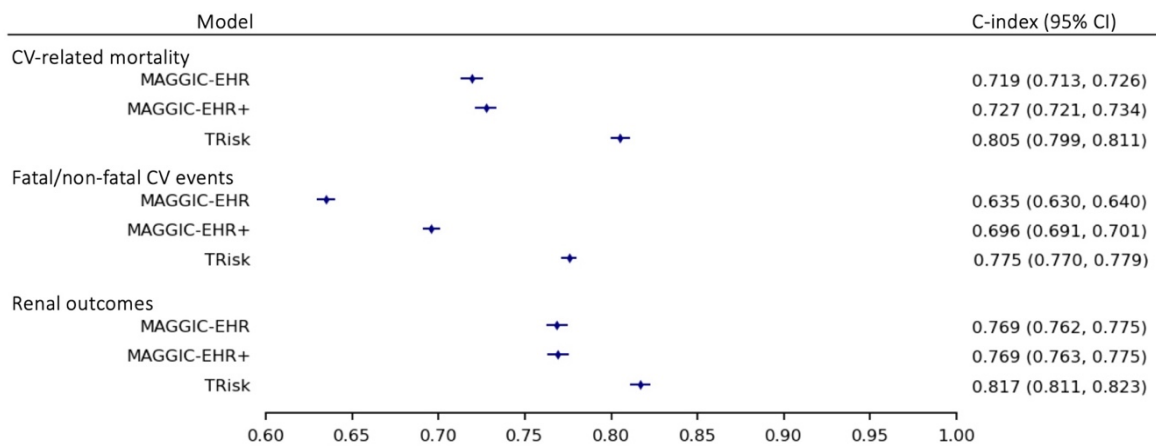

*Figure S11. Discriminative performance of models for 36-month risk prediction of various outcomes on UK validation data.*
*Discrimination is provided in this forest plot as assessed by C-index with 95% confidence intervals (CI) for various outcome investigations at 36-month timepoint; CV: cardiovascular.*

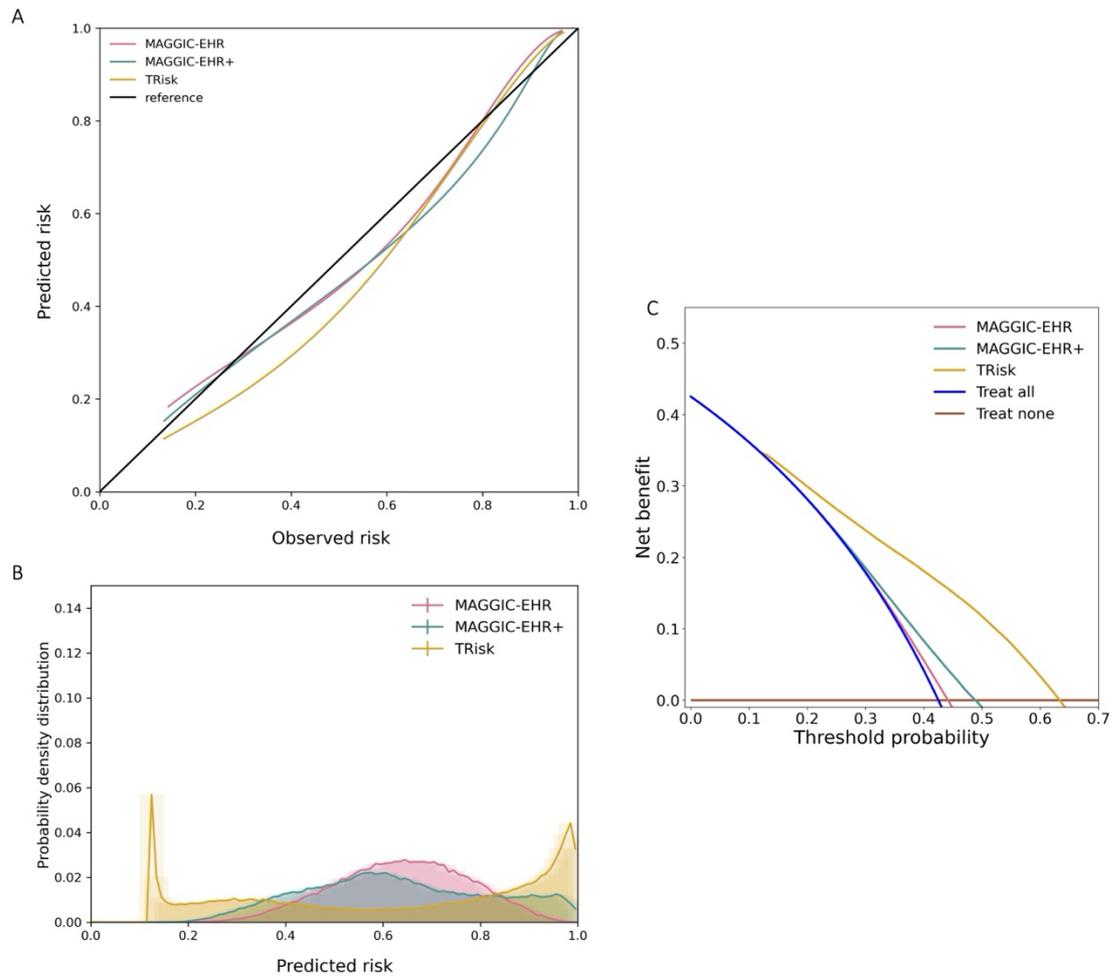

*Figure S12. Calibration curves, distribution of predicted risk of models, decision curve analyses 36-month non-fatal/fatal cardiovascular event prediction on UK validation data.*
*(A) Calibration curves, (B) distribution of predicted risk, and (C) decision curve analyses are presented for all models. Decision curve analysis (including censored observations) has been conducted for all models. Threshold probability is shown on the x-axis and the net benefit, a function of threshold probability, is shown on the y-axis and is the difference between the proportion of true positives and false positives weighted by odds of the respective decision threshold.*

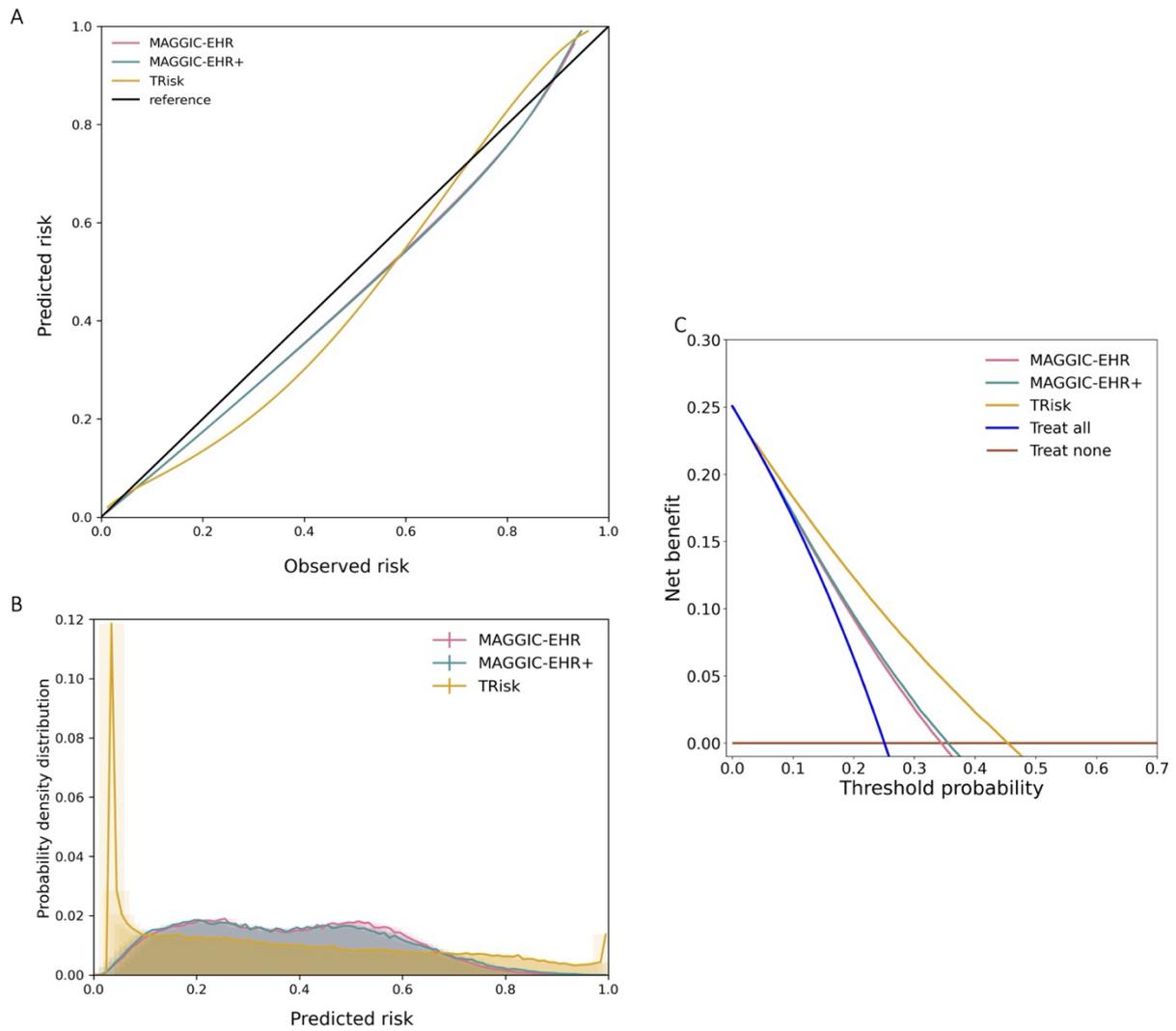

*Figure S13. Calibration curves, distribution of predicted risk of models, decision curve analyses 36-month CV-related mortality prediction on UK validation data.*
*(A) Calibration curves, (B) distribution of predicted risk, and (C) decision curve analyses are presented for all models. Decision curve analysis (including censored observations) has been conducted for all models. Threshold probability is shown on the x-axis and the net benefit, a function of threshold probability, is shown on the y-axis and is the difference between the proportion of true positives and false positives weighted by odds of the respective decision threshold.*

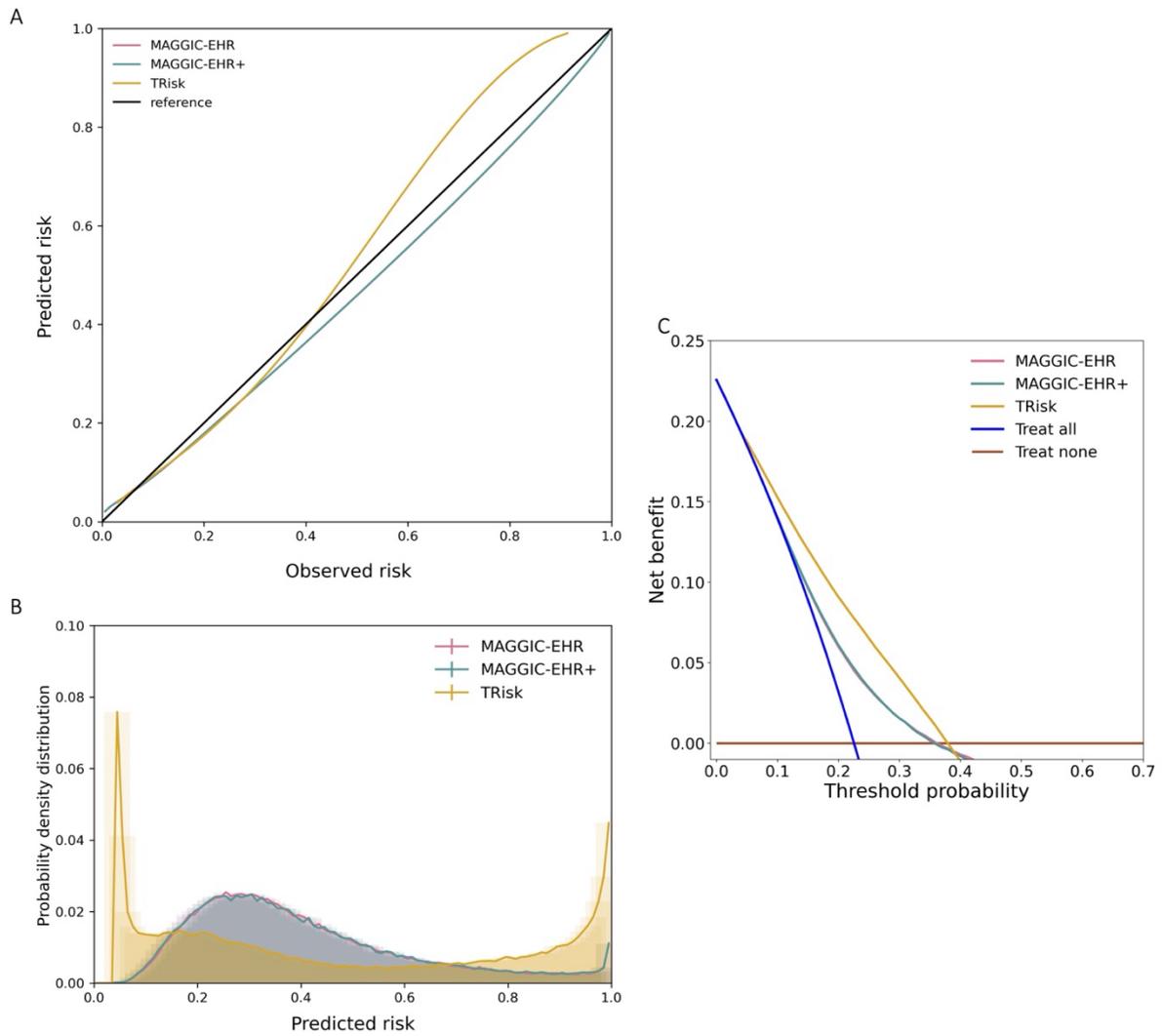

*Figure S14. Calibration curves, distribution of predicted risk of models, decision curve analyses 36-month renal outcomes prediction on UK validation data.*
*(A) Calibration curves, (B) distribution of predicted risk, and (C) decision curve analyses are presented for all models. Decision curve analysis (including censored observations) has been conducted for all models. Threshold probability is shown on the x-axis and the net benefit, a function of threshold probability, is shown on the y-axis and is the difference between the proportion of true positives and false positives weighted by odds of the respective decision threshold.*

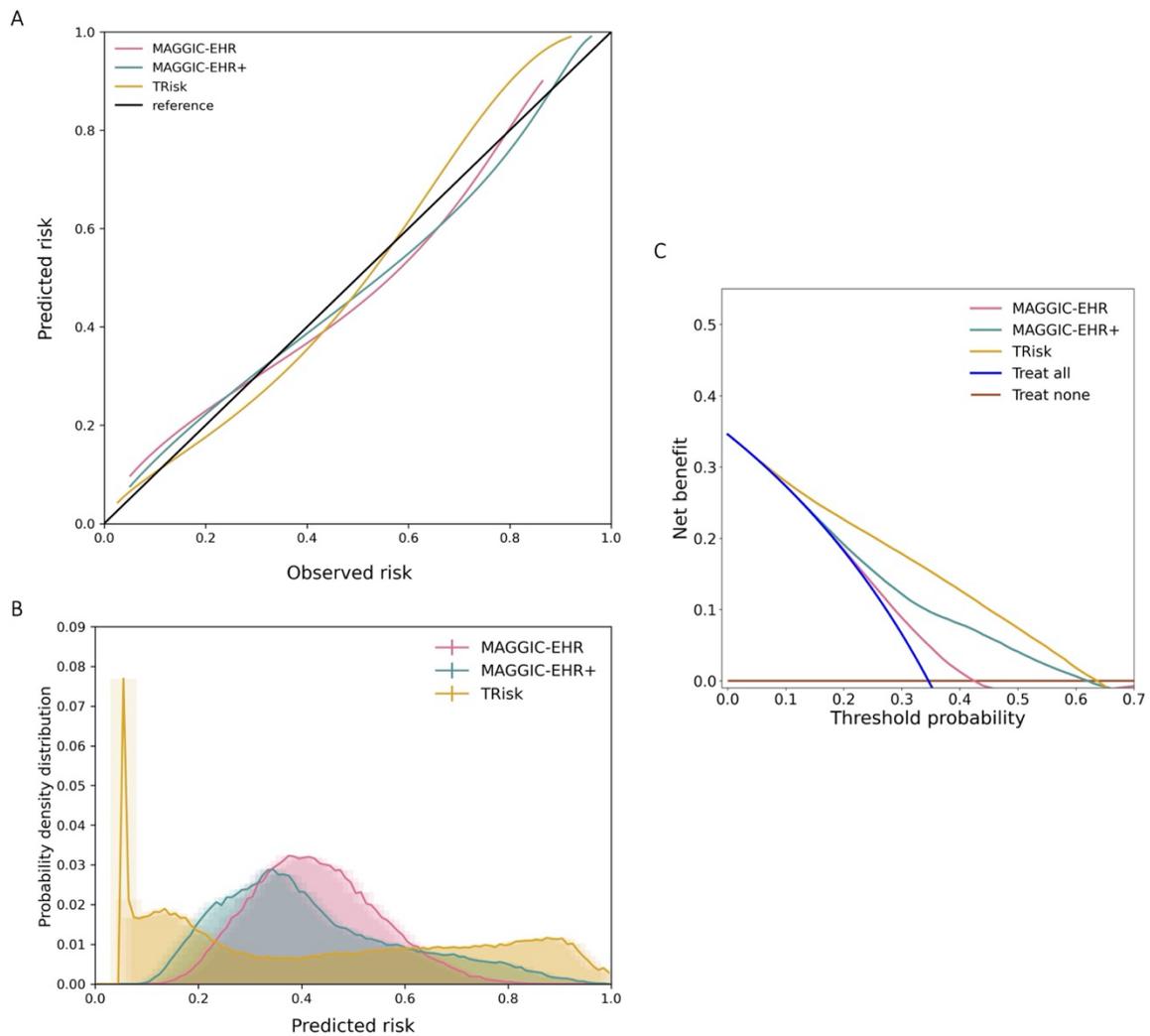

*Figure S15. Calibration curves, distribution of predicted risk of models, decision curve analyses 12-month non-fatal/fatal cardiovascular event prediction on UK validation data.*
*(A) Calibration curves, (B) distribution of predicted risk, and (C) decision curve analyses are presented for all models. Decision curve analysis (including censored observations) has been conducted for all models. Threshold probability is shown on the x-axis and the net benefit, a function of threshold probability, is shown on the y-axis and is the difference between the proportion of true positives and false positives weighted by odds of the respective decision threshold.*

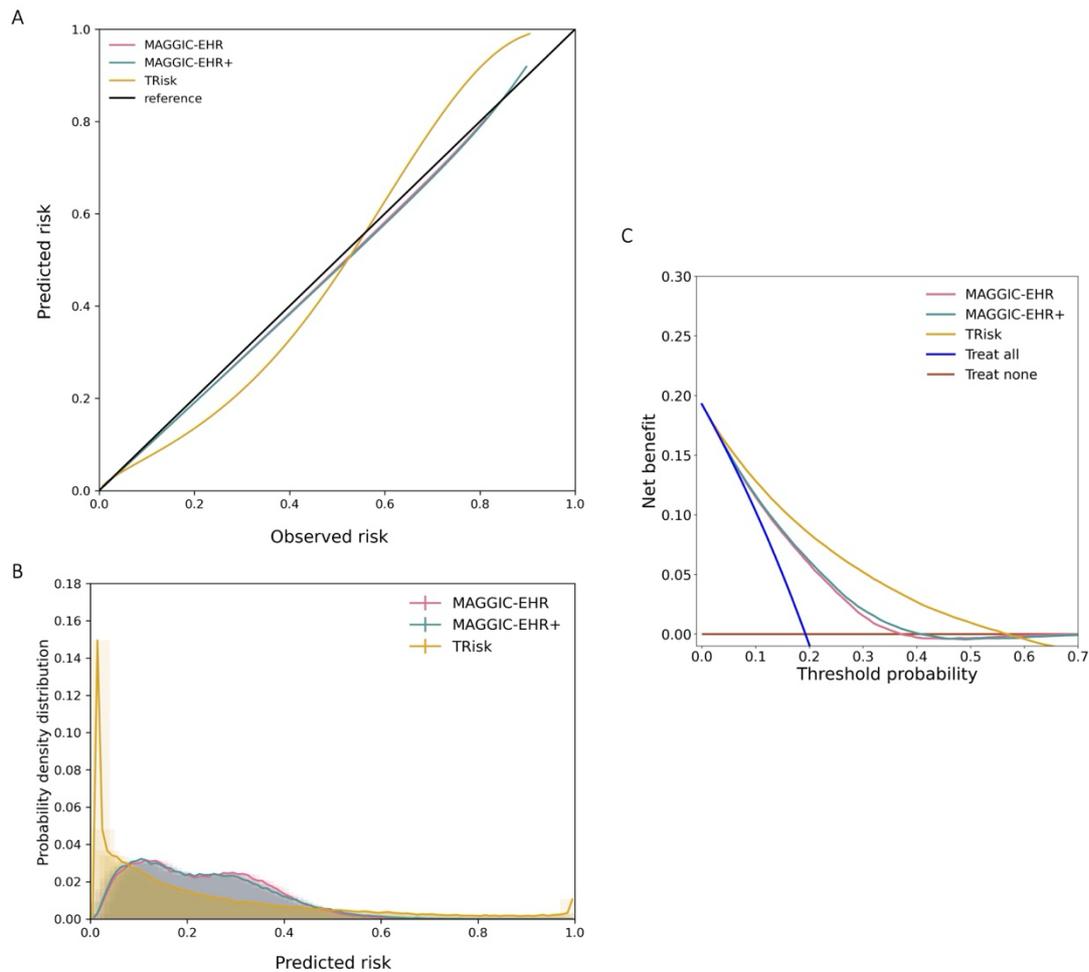

*Figure S16. Calibration curves, distribution of predicted risk of models, decision curve analyses 12-month CV-related mortality prediction on UK validation data.*
*(A) Calibration curves, (B) distribution of predicted risk, and (C) decision curve analyses are presented for all models. Decision curve analysis (including censored observations) has been conducted for all models. Threshold probability is shown on the x-axis and the net benefit, a function of threshold probability, is shown on the y-axis and is the difference between the proportion of true positives and false positives weighted by odds of the respective decision threshold.*

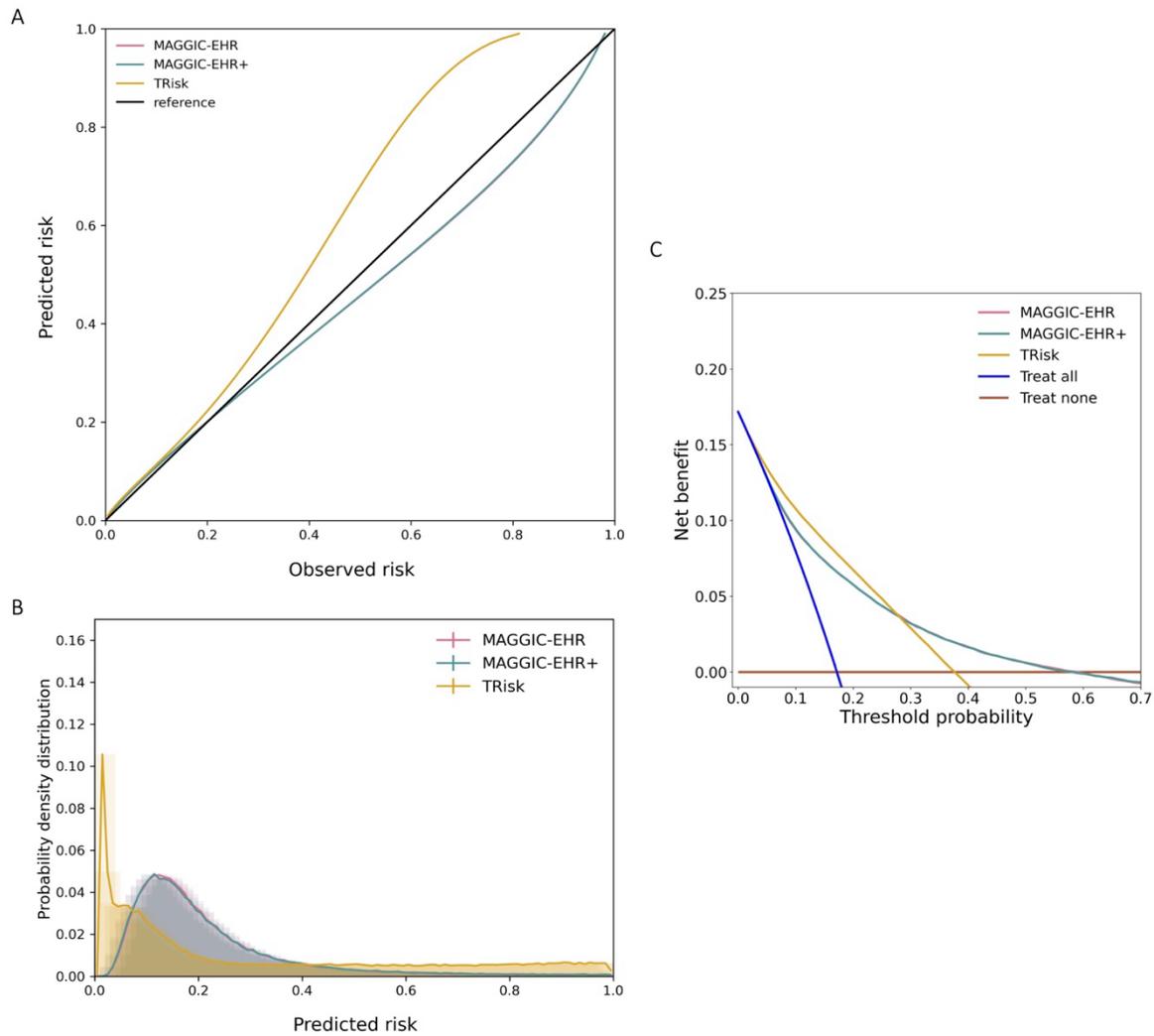

*Figure S17. Calibration curves, distribution of predicted risk of models, decision curve analyses 12-month renal outcomes prediction on UK validation data.*
*(A) Calibration curves, (B) distribution of predicted risk, and (C) decision curve analyses are presented for all models. Decision curve analysis (including censored observations) has been conducted for all models. Threshold probability is shown on the x-axis and the net benefit, a function of threshold probability, is shown on the y-axis and is the difference between the proportion of true positives and false positives weighted by odds of the respective decision threshold.*

Supplementary Tables

*Table S1. Model hyperparameters and other settings for TRisk*

| Hyperparameters | Value |
|---|---|
| Number of layers | 6 |
| Maximum sequence length | 512 |
| Hidden size | 150 |
| Hidden dropout rate | 0.3 |
| Attention dropout rate | 0.4 |
| Number of attention heads | 6 |
| Intermediate size | 108 |
| Pooling layer size | 150 |
| Hidden activation function | Gaussian Error Linear Unit (GELU) function |
| Learning rate | 0.00008 |
| Weight decay | 0.02 |
| Warmup proportion (Adam optimiser hyperparameter) | 0.1 |
| Embedding concatenation | True |
| Embedding non-linear transformation | Hyperbolic tangent function |
| $\lambda$ weightage for explicit calibration-related modelling | 2.0 |
| Interpolation for explicit calibration-related modelling | True |
| Time bins for explicit calibration-related modelling | 48 (i.e., 1 per month) |

*Table S2. Area under the precision-recall curve (AUPRC) metrics for all models for all 36-month risk prediction investigations on UK validation data.*

|  |  | AUPRC |
|---|---|---|
| All-cause mortality | MAGGIC-EHR | 0.623 |
|  | MAGGIC-EHR+ | 0.633 |
|  | TRisk | 0.797 |
| Non-fatal/fatal cardiovascular events | MAGGIC-EHR | 0.510 |
|  | MAGGIC-EHR+ | 0.616 |
|  | TRisk | 0.700 |
| Cardiovascular-related mortality | MAGGIC-EHR | 0.388 |
|  | MAGGIC-EHR+ | 0.403 |
|  | TRisk | 0.524 |
| Renal outcomes | MAGGIC-EHR | 0.478 |
|  | MAGGIC-EHR+ | 0.478 |
|  | TRisk | 0.529 |

*AUPRC: area under precision-recall curve*

*Table S2. Area under the precision-recall curve (AUPRC) metrics for all models for all 36-month risk prediction investigations on UK validation data.*

*Table S3. Subgroup discrimination analysis for 36-month all-cause mortality prediction investigations on UK validation data*

| Analysis | Model | Subgroup | Count | Concordance index (95% CI) | AUPRC |
|---|---|---|---|---|---|
| Sex | MAGGIC-EHR | Male | 53,938 | 0.741 (0.734, 0.748) | 0.634 |
| | MAGGIC-EHR | Female | 45,444 | 0.712 (0.705, 0.719) | 0.609 |
| | MAGGIC-EHR+ | Male | 53,938 | 0.747 (0.741, 0.754) | 0.643 |
| | MAGGIC-EHR+ | Female | 45,444 | 0.719 (0.711, 0.726) | 0.619 |
| | TRisk | Male | 53,938 | 0.849 (0.843, 0.854) | 0.801 |
| | TRisk | Female | 45,444 | 0.841 (0.835, 0.847) | 0.794 |
| Age | MAGGIC-EHR | <60 years | 15,738 | 0.724 (0.710, 0.737) | 0.570 |
| | MAGGIC-EHR | ≥60 years | 83,644 | 0.729 (0.724, 0.734) | 0.633 |
| | MAGGIC-EHR+ | <60 years | 15,738 | 0.733 (0.719, 0.746) | 0.580 |
| | MAGGIC-EHR+ | ≥60 years | 83,644 | 0.735 (0.730, 0.740) | 0.642 |
| | TRisk | <60 years | 15,738 | 0.845 (0.834, 0.856) | 0.745 |
| | TRisk | ≥60 years | 83,644 | 0.844 (0.840, 0.848) | 0.804 |
| Baseline systolic blood pressure | MAGGIC-EHR | <130 mmHg | 37,798 | 0.744 (0.736, 0.752) | 0.632 |
| | MAGGIC-EHR | ≥130 mmHg | 61,584 | 0.718 (0.712, 0.724) | 0.617 |
| | MAGGIC-EHR+ | <130 mmHg | 37,798 | 0.752 (0.744, 0.760) | 0.644 |
| | MAGGIC-EHR+ | ≥130 mmHg | 61,584 | 0.724 (0.717, 0.730) | 0.625 |
| | TRisk | <130 mmHg | 37,798 | 0.848 (0.842, 0.855) | 0.795 |
| | TRisk | ≥130 mmHg | 61,584 | 0.843 (0.838, 0.848) | 0.799 |
| HF subtype | MAGGIC-EHR | Preserved | 4,350 | 0.662 (0.623, 0.700) | 0.271 |
| | MAGGIC-EHR | Reduced | 21,468 | 0.681 (0.667, 0.695) | 0.366 |
| | MAGGIC-EHR+ | Preserved | 4,350 | 0.674 (0.636, 0.712) | 0.285 |
| | MAGGIC-EHR+ | Reduced | 21,468 | 0.690 (0.676, 0.704) | 0.378 |
| | TRisk | Preserved | 4,350 | 0.818 (0.787, 0.850) | 0.511 |
| | TRisk | Reduced | 21,468 | 0.799 (0.786, 0.811) | 0.575 |
| Baseline disease | MAGGIC-EHR | Diabetes | 29,659 | 0.709 (0.700, 0.718) | 0.607 |
| | MAGGIC-EHR+ | Diabetes | 29,659 | 0.715 (0.706, 0.723) | 0.613 |
| | TRisk | Diabetes | 29,659 | 0.829 (0.822, 0.837) | 0.785 |
| | MAGGIC-EHR | Atrial fibrillation | 48,033 | 0.713 (0.706, 0.720) | 0.629 |
| | MAGGIC-EHR+ | Atrial fibrillation | 48,033 | 0.718 (0.711, 0.725) | 0.636 |
| | TRisk | Atrial fibrillation | 48,033 | 0.833 (0.827, 0.838) | 0.797 |
| | MAGGIC-EHR | Myocardial infarction | 28,500 | 0.729 (0.719, 0.738) | 0.599 |
| | MAGGIC-EHR+ | Myocardial infarction | 28,500 | 0.738 (0.728, 0.747) | 0.611 |
| | TRisk | Myocardial infarction | 28,500 | 0.830 (0.822, 0.838) | 0.760 |
| Baseline medication | MAGGIC-EHR | Beta-blockers | 34,258 | 0.715 (0.707, 0.724) | 0.615 |
| | MAGGIC-EHR+ | Beta-blockers | 34,258 | 0.721 (0.712, 0.729) | 0.623 |
| | TRisk | Beta-blockers | 34,258 | 0.828 (0.821, 0.835) | 0.782 |
| | MAGGIC-EHR | Angiotensin-converting-enzyme inhibitors | 52,799 | 0.726 (0.718, 0.733) | 0.568 |
| | MAGGIC-EHR+ | Angiotensin-converting-enzyme inhibitors | 52,799 | 0.734 (0.727, 0.741) | 0.584 |
| | TRisk | Angiotensin-converting-enzyme inhibitors | 52,799 | 0.834 (0.828, 0.840) | 0.751 |
| | MAGGIC-EHR | Angiotensin receptor blockers | 26,205 | 0.720 (0.709, 0.730) | 0.559 |
| | MAGGIC-EHR+ | Angiotensin receptor blockers | 26,205 | 0.726 (0.716, 0.736) | 0.570 |
| | TRisk | Angiotensin receptor blockers | 26,205 | 0.837 (0.829, 0.846) | 0.751 |

*HF: heart failure; AUPRC: area under precision-recall curve; CI: confidence interval*

*Table S4. Integrated calibration index (ICI) for various 36-month risk prediction investigations across all models on UK validation data*

| | Integrated Calibration Index (ICI)* | | | |
|---|---|---|---|---|
| | All-cause mortality | Non-fatal/fatal cardiovascular events | Cardiovascular-related mortality | Renal outcomes |
| MAGGIC-EHR | 0.0478 | 0.0433 | 0.0102 | 0.0381 |
| MAGGIC-EHR+ | 0.0496 | 0.0543 | 0.0110 | 0.0380 |
| TRisk | 0.0419 | 0.0503 | 0.0501 | 0.0481 |

*lower is better*

*Table S5. Integrated calibration index (ICI) for various outcome 12-month risk prediction investigations across all models on UK validation data*

| | Integrated Calibration Index (ICI)* | | | |
|---|---|---|---|---|
| | All-cause mortality | Non-fatal/fatal cardiovascular events | Cardiovascular-related mortality | Renal outcomes |
| MAGGIC-EHR | 0.0527 | 0.0432 | 0.0102 | 0.0130 |
| MAGGIC-EHR+ | 0.0510 | 0.0254 | 0.0110 | 0.0128 |
| TRisk | 0.0386 | 0.0387 | 0.0431 | 0.0598 |

*lower is better*

*Table S6. Impact analyses at the 50% decision threshold for 12- and 36-month all-cause mortality prediction on UK validation data*

| Timepoint | Models | Metrics | | | | |
|---|---|---|---|---|---|---|
| | | PP | FP | FN | PPV | Sensitivity |
| 12-month | MAGGIC-EHR | 22,924 | 9,882 | 18,592 | 0.569 | 0.412 |
| | TRisk | 30,769 | 9,183 | 10,048 | 0.702 | 0.682 |
| 36-month | MAGGIC-EHR | 56,158 | 24,871 | 9,770 | 0.557 | 0.762 |
| | TRisk | 50,848 | 17,178 | 7,387 | 0.662 | 0.821 |

PP: predicted positive; FP: false positives; FN: false negatives; PPV: positive predictive value

*Table S7. Population characteristics of the MIMIC-IV cohort*

|  | Fine-tuning: 8,707 patients (40%) | Validation: 13,060 patients (60%) |
|---|---|---|
| Women (%) | 4,789 (55.0) | 7,211 (55.2) |
| Median age in years (IQI) | 74 (65, 83) | 74 (65, 82) |
| Smoking status | | |
|     Current/ex-smoker (%) $^{\Omega}$ | 3,891 (44.7) | 5,853 (44.8) |
|     Non-smoker (%) | 4,816 (55.3) | 7,207 (55.2) |
| Median creatinine (µmol/L) (IQI) $^{\dagger}$ | 88.4 (79.6, 106.1) | 88.4 (79.6, 106.1) |
| Median sodium (mmol/L) (IQI) $^{\dagger}$ | 138.8 (136.8, 140.7) | 138.8 (136.7, 140.7) |
| Median BMI (kg/m$^2$) (IQI) $^{\dagger}$ | 28.3 (24.4, 33.2) | 28.4 (24.3, 33.5) |
| Median SBP (mmHg) (IQI) $^{\dagger}$ | 129.0 (119.2, 139.7) | 128.9 (119.1, 139.3) |
| <18 months after HF (%) | 7,158 (82.2) | 10,749 (82.3) |
| Diseases at baseline | | |
|     Diabetes | 3,369 (38.7) | 4,984 (38.2) |
|     COPD | 2,280 (26.2) | 3,323 (25.4) |
|     Atrial fibrillation | 4,124 (47.4) | 6,131 (46.9) |
|     Stroke | 568 (6.5) | 877 (6.7) |
| Medication use at baseline | | |
|     Beta blockers | 6,389 (73.4) | 9,678 (74.1) |

%: percent; IQI: interquartile interval; BMI: body mass index; SBP: systolic blood pressure; COPD: Chronic obstructive pulmonary disease; AF: atrial fibrillation; $^{\Omega}$ current/ex-smoker status identified by F17 and Z87.891 ICD-10 codes; $^{\dagger}$indicates missing variables; BMI (41.7% missingness), SBP (44.1%), creatinine (4.5%), sodium (4.7%).

Table S8. *Area under the precision-recall curve (AUPRC) metrics for all models for all 12- and 36-month risk prediction of all-cause mortality prediction on MIMIC-IV validation data*

| | AUPRC | |
|---|---|---|
| Model | 12-month | 36-month |
| TRisk (MIMIC-IV derived) | 0.403 | 0.361 |
| TRisk (CPRD derived) | 0.541 | 0.551 |
| TRisk (using transfer learning) | 0.690 | 0.693 |

Table S8. *Area under the precision-recall curve (AUPRC) metrics for all models for all 12- and 36-month risk prediction of all-cause mortality prediction on MIMIC-IV validation data*

*Table S9. Integrated calibration index (ICI) for 12- and 36-month risk prediction of all-cause mortality on MIMIC-IV validation data across all models*

| | Integrated calibration index (ICI)* | |
|---|---|---|
| Model | 12-month | 36-month |
| TRisk (MIMIC-IV derived) | 0.088 | 0.068 |
| TRisk (CPRD derived) | 0.312 | 0.166 |
| TRisk (using transfer learning) | 0.051 | 0.053 |

*lower is better*

*Table S10. Subgroup discrimination analysis for 36-month cardiovascular-related mortality prediction investigations on UK validation data*

| Analysis | Model | Subgroup | Count | Concordance index (95% CI) | AUPRC |
|---|---|---|---|---|---|
| Sex | MAGGIC-EHR | Male | 53,938 | 0.729 (0.720, 0.737) | 0.403 |
| | MAGGIC-EHR | Female | 45,444 | 0.708 (0.698, 0.717) | 0.366 |
| | MAGGIC-EHR+ | Male | 53,938 | 0.735 (0.726, 0.743) | 0.416 |
| | MAGGIC-EHR+ | Female | 45,444 | 0.718 (0.709, 0.728) | 0.386 |
| | TRisk | Male | 53,938 | 0.810 (0.802, 0.817) | 0.544 |
| | TRisk | Female | 45,444 | 0.800 (0.792, 0.809) | 0.518 |
| Age | MAGGIC-EHR | <60 years | 15,738 | 0.725 (0.707, 0.743) | 0.352 |
| | MAGGIC-EHR | ≥60 years | 83,644 | 0.719 (0.712, 0.725) | 0.396 |
| | MAGGIC-EHR+ | <60 years | 15,738 | 0.734 (0.716, 0.752) | 0.364 |
| | MAGGIC-EHR+ | ≥60 years | 83,644 | 0.726 (0.720, 0.733) | 0.411 |
| | TRisk | <60 years | 15,738 | 0.802 (0.786, 0.818) | 0.471 |
| | TRisk | ≥60 years | 83,644 | 0.804 (0.798, 0.810) | 0.540 |
| Baseline systolic blood pressure | MAGGIC-EHR | <130 mmHg | 37,798 | 0.734 (0.724, 0.744) | 0.384 |
| | MAGGIC-EHR | ≥130 mmHg | 61,584 | 0.710 (0.702, 0.718) | 0.390 |
| | MAGGIC-EHR+ | <130 mmHg | 37,798 | 0.742 (0.732, 0.753) | 0.401 |
| | MAGGIC-EHR+ | ≥130 mmHg | 61,584 | 0.718 (0.710, 0.726) | 0.404 |
| | TRisk | <130 mmHg | 37,798 | 0.806 (0.797, 0.816) | 0.509 |
| | TRisk | ≥130 mmHg | 61,584 | 0.804 (0.797, 0.811) | 0.544 |
| HF subtype | MAGGIC-EHR | Preserved | 4,350 | 0.665 (0.613, 0.717) | 0.156 |
| | MAGGIC-EHR | Reduced | 21,468 | 0.688 (0.671, 0.705) | 0.261 |
| | MAGGIC-EHR+ | Preserved | 4,350 | 0.687 (0.636, 0.738) | 0.171 |
| | MAGGIC-EHR+ | Reduced | 21,468 | 0.698 (0.681, 0.715) | 0.270 |
| | TRisk | Preserved | 4,350 | 0.806 (0.762, 0.850) | 0.289 |
| | TRisk | Reduced | 21,468 | 0.777 (0.761, 0.792) | 0.392 |
| Baseline disease | MAGGIC-EHR | Diabetes | 29,659 | 0.702 (0.691, 0.713) | 0.406 |
| | MAGGIC-EHR+ | Diabetes | 29,659 | 0.706 (0.695, 0.717) | 0.415 |
| | TRisk | Diabetes | 29,659 | 0.789 (0.779, 0.799) | 0.550 |
| | MAGGIC-EHR | Atrial fibrillation | 48,033 | 0.701 (0.692, 0.709) | 0.404 |
| | MAGGIC-EHR+ | Atrial fibrillation | 48,033 | 0.708 (0.700, 0.717) | 0.417 |
| | TRisk | Atrial fibrillation | 48,033 | 0.789 (0.781, 0.797) | 0.539 |
| | MAGGIC-EHR | Myocardial infarction | 28,500 | 0.724 (0.713, 0.735) | 0.443 |
| | MAGGIC-EHR+ | Myocardial infarction | 28,500 | 0.730 (0.720, 0.741) | 0.446 |
| | TRisk | Myocardial infarction | 28,500 | 0.797 (0.787, 0.807) | 0.551 |
| Baseline medication | MAGGIC-EHR | Beta-blockers | 34,258 | 0.712 (0.701, 0.722) | 0.417 |
| | MAGGIC-EHR+ | Beta-blockers | 34,258 | 0.719 (0.709, 0.729) | 0.431 |
| | TRisk | Beta-blockers | 34,258 | 0.791 (0.782, 0.800) | 0.544 |
| | MAGGIC-EHR | Angiotensin-converting-enzyme inhibitors | 52,799 | 0.723 (0.714, 0.732) | 0.375 |
| | MAGGIC-EHR+ | Angiotensin-converting-enzyme inhibitors | 52,799 | 0.732 (0.723, 0.741) | 0.392 |
| | TRisk | Angiotensin-converting-enzyme inhibitors | 52,799 | 0.800 (0.792, 0.808) | 0.510 |
| | MAGGIC-EHR | Angiotensin receptor blockers | 26,205 | 0.717 (0.705, 0.730) | 0.374 |
| | MAGGIC-EHR+ | Angiotensin receptor blockers | 26,205 | 0.727 (0.715, 0.740) | 0.392 |
| | TRisk | Angiotensin receptor blockers | 26,205 | 0.803 (0.792, 0.814) | 0.513 |

*HF: heart failure; AUPRC: area under precision-recall curve; CI: confidence interval*

*Table S11. Subgroup discrimination analysis for 36-month non-fatal/fatal cardiovascular event prediction investigations on UK validation data*

| Analysis | Model | Subgroup | Count | Concordance index (95% CI) | AUPRC |
|---|---|---|---|---|---|
| Sex | MAGGIC-EHR | Male | 53,938 | 0.631 (0.624, 0.638) | 0.539 |
| | MAGGIC-EHR | Female | 45,444 | 0.630 (0.623, 0.639) | 0.448 |
| | MAGGIC-EHR+ | Male | 53,938 | 0.686 (0.680, 0.693) | 0.641 |
| | MAGGIC-EHR+ | Female | 45,444 | 0.702 (0.694, 0.710) | 0.572 |
| | TRisk | Male | 53,938 | 0.763 (0.757, 0.769) | 0.715 |
| | TRisk | Female | 45,444 | 0.786 (0.779, 0.793) | 0.677 |
| Age | MAGGIC-EHR | <60 years | 15,738 | 0.631 (0.617, 0.645) | 0.471 |
| | MAGGIC-EHR | ≥60 years | 83,644 | 0.633 (0.627, 0.638) | 0.515 |
| | MAGGIC-EHR+ | <60 years | 15,738 | 0.712 (0.699, 0.726) | 0.592 |
| | MAGGIC-EHR+ | ≥60 years | 83,644 | 0.691 (0.686, 0.697) | 0.619 |
| | TRisk | <60 years | 15,738 | 0.789 (0.777, 0.801) | 0.679 |
| | TRisk | ≥60 years | 83,644 | 0.771 (0.766, 0.776) | 0.703 |
| Baseline systolic blood pressure | MAGGIC-EHR | <130 mmHg | 37,798 | 0.640 (0.632, 0.649) | 0.515 |
| | MAGGIC-EHR | ≥130 mmHg | 61,584 | 0.632 (0.625, 0.638) | 0.507 |
| | MAGGIC-EHR+ | <130 mmHg | 37,798 | 0.700 (0.692, 0.708) | 0.617 |
| | MAGGIC-EHR+ | ≥130 mmHg | 61,584 | 0.694 (0.687, 0.700) | 0.615 |
| | TRisk | <130 mmHg | 37,798 | 0.772 (0.764, 0.779) | 0.689 |
| | TRisk | ≥130 mmHg | 61,584 | 0.777 (0.771, 0.782) | 0.706 |
| HF subtype | MAGGIC-EHR | Preserved | 4,350 | 0.634 (0.605, 0.664) | 0.424 |
| | MAGGIC-EHR | Reduced | 21,468 | 0.622 (0.611, 0.634) | 0.534 |
| | MAGGIC-EHR+ | Preserved | 4,350 | 0.708 (0.681, 0.736) | 0.532 |
| | MAGGIC-EHR+ | Reduced | 21,468 | 0.693 (0.683, 0.704) | 0.625 |
| | TRisk | Preserved | 4,350 | 0.792 (0.767, 0.817) | 0.655 |
| | TRisk | Reduced | 21,468 | 0.747 (0.737, 0.757) | 0.707 |
| Baseline disease | MAGGIC-EHR | Diabetes | 29,659 | 0.620 (0.611, 0.629) | 0.554 |
| | MAGGIC-EHR+ | Diabetes | 29,659 | 0.665 (0.657, 0.674) | 0.640 |
| | TRisk | Diabetes | 29,659 | 0.754 (0.746, 0.762) | 0.731 |
| | MAGGIC-EHR | Atrial fibrillation | 48,033 | 0.630 (0.622, 0.637) | 0.512 |
| | MAGGIC-EHR+ | Atrial fibrillation | 48,033 | 0.691 (0.684, 0.698) | 0.612 |
| | TRisk | Atrial fibrillation | 48,033 | 0.770 (0.763, 0.776) | 0.695 |
| | MAGGIC-EHR | Myocardial infarction | 28,500 | 0.597 (0.589, 0.605) | 0.700 |
| | MAGGIC-EHR+ | Myocardial infarction | 28,500 | 0.603 (0.595, 0.611) | 0.704 |
| | TRisk | Myocardial infarction | 28,500 | 0.682 (0.674, 0.689) | 0.766 |
| Baseline medication | MAGGIC-EHR | Beta-blockers | 34,258 | 0.629 (0.621, 0.637) | 0.568 |
| | MAGGIC-EHR+ | Beta-blockers | 34,258 | 0.683 (0.675, 0.691) | 0.672 |
| | TRisk | Beta-blockers | 34,258 | 0.756 (0.749, 0.764) | 0.736 |
| | MAGGIC-EHR | Angiotensin-converting-enzyme inhibitors | 52,799 | 0.633 (0.626, 0.640) | 0.539 |
| | MAGGIC-EHR+ | Angiotensin-converting-enzyme inhibitors | 52,799 | 0.694 (0.687, 0.700) | 0.643 |
| | TRisk | Angiotensin-converting-enzyme inhibitors | 52,799 | 0.764 (0.758, 0.770) | 0.722 |
| | MAGGIC-EHR | Angiotensin receptor blockers | 26,205 | 0.632 (0.623, 0.642) | 0.528 |
| | MAGGIC-EHR+ | Angiotensin receptor blockers | 26,205 | 0.696 (0.687, 0.706) | 0.638 |
| | TRisk | Angiotensin receptor blockers | 26,205 | 0.767 (0.758, 0.776) | 0.719 |

*HF: heart failure; AUPRC: area under precision-recall curve; CI: confidence interval*

*Table S12. Subgroup discrimination analysis for 36-month renal outcomes prediction investigations on UK validation data*

| Analysis | Model | Subgroup | Count | Concordance index (95% CI) | AUPRC |
|---|---|---|---|---|---|
| Sex | MAGGIC-EHR | Male | 53,938 | 0.782 (0.774, 0.791) | 0.496 |
| | MAGGIC-EHR | Female | 45,444 | 0.751 (0.741, 0.76) | 0.455 |
| | MAGGIC-EHR+ | Male | 53,938 | 0.783 (0.774, 0.791) | 0.497 |
| | MAGGIC-EHR+ | Female | 45,444 | 0.751 (0.742, 0.761) | 0.454 |
| | TRisk | Male | 53,938 | 0.829 (0.821, 0.836) | 0.545 |
| | TRisk | Female | 45,444 | 0.802 (0.793, 0.811) | 0.504 |
| Age | MAGGIC-EHR | <60 years | 15,738 | 0.787 (0.770, 0.804) | 0.480 |
| | MAGGIC-EHR | ≥60 years | 83,644 | 0.765 (0.758, 0.771) | 0.477 |
| | MAGGIC-EHR+ | <60 years | 15,738 | 0.788 (0.772, 0.805) | 0.480 |
| | MAGGIC-EHR+ | ≥60 years | 83,644 | 0.765 (0.758, 0.772) | 0.477 |
| | TRisk | <60 years | 15,738 | 0.851 (0.836, 0.865) | 0.556 |
| | TRisk | ≥60 years | 83,644 | 0.809 (0.803, 0.816) | 0.522 |
| Baseline systolic blood pressure | MAGGIC-EHR | <130 mmHg | 37,798 | 0.775 (0.764, 0.785) | 0.455 |
| | MAGGIC-EHR | ≥130 mmHg | 61,584 | 0.765 (0.757, 0.772) | 0.489 |
| | MAGGIC-EHR+ | <130 mmHg | 37,798 | 0.775 (0.765, 0.786) | 0.456 |
| | MAGGIC-EHR+ | ≥130 mmHg | 61,584 | 0.765 (0.757, 0.773) | 0.489 |
| | TRisk | <130 mmHg | 37,798 | 0.820 (0.810, 0.830) | 0.503 |
| | TRisk | ≥130 mmHg | 61,584 | 0.815 (0.807, 0.822) | 0.538 |
| HF subtype | MAGGIC-EHR | Preserved | 4,350 | 0.780 (0.751, 0.809) | 0.530 |
| | MAGGIC-EHR | Reduced | 21,468 | 0.777 (0.764, 0.790) | 0.535 |
| | MAGGIC-EHR+ | Preserved | 4,350 | 0.781 (0.752, 0.810) | 0.533 |
| | MAGGIC-EHR+ | Reduced | 21,468 | 0.776 (0.763, 0.789) | 0.536 |
| | TRisk | Preserved | 4,350 | 0.818 (0.791, 0.845) | 0.590 |
| | TRisk | Reduced | 21,468 | 0.807 (0.795, 0.820) | 0.573 |
| Baseline disease | MAGGIC-EHR | Diabetes | 29,659 | 0.743 (0.733, 0.753) | 0.546 |
| | MAGGIC-EHR+ | Diabetes | 29,659 | 0.743 (0.732, 0.753) | 0.546 |
| | TRisk | Diabetes | 29,659 | 0.803 (0.794, 0.812) | 0.598 |
| | MAGGIC-EHR | Atrial fibrillation | 48,033 | 0.751 (0.742, 0.760) | 0.486 |
| | MAGGIC-EHR+ | Atrial fibrillation | 48,033 | 0.751 (0.742, 0.760) | 0.486 |
| | TRisk | Atrial fibrillation | 48,033 | 0.800 (0.792, 0.808) | 0.524 |
| | MAGGIC-EHR | Myocardial infarction | 28,500 | 0.766 (0.755, 0.777) | 0.522 |
| | MAGGIC-EHR+ | Myocardial infarction | 28,500 | 0.768 (0.757, 0.779) | 0.520 |
| | TRisk | Myocardial infarction | 28,500 | 0.822 (0.812, 0.832) | 0.583 |
| Baseline medication | MAGGIC-EHR | Beta-blockers | 34,258 | 0.751 (0.741, 0.761) | 0.509 |
| | MAGGIC-EHR+ | Beta-blockers | 34,258 | 0.753 (0.743, 0.762) | 0.511 |
| | TRisk | Beta-blockers | 34,258 | 0.801 (0.792, 0.810) | 0.564 |
| | MAGGIC-EHR | Angiotensin-converting-enzyme inhibitors | 52,799 | 0.769 (0.761, 0.777) | 0.502 |
| | MAGGIC-EHR+ | Angiotensin-converting-enzyme inhibitors | 52,799 | 0.769 (0.761, 0.777) | 0.502 |
| | TRisk | Angiotensin-converting-enzyme inhibitors | 52,799 | 0.811 (0.804, 0.819) | 0.553 |
| | MAGGIC-EHR | Angiotensin receptor blockers | 26,205 | 0.751 (0.740, 0.762) | 0.544 |
| | MAGGIC-EHR+ | Angiotensin receptor blockers | 26,205 | 0.752 (0.741, 0.763) | 0.542 |
| | TRisk | Angiotensin receptor blockers | 26,205 | 0.805 (0.795, 0.815) | 0.600 |

*HF: heart failure; AUPRC: area under precision-recall curve; CI: confidence interval*

*Table S13. Multivariable risk models for patients with heart failure*

| Authors | Risk prediction model | Derivation cohort | | | | | # Patients; Setting |
|---|---|---|---|---|---|---|---|
| | | Sample size | Country/ region | Patient population | Outcomes investigated | Variables included | |
| Levy et al. (2006) [15] | Seattle Heart Failure Model | 1,125 | USA and Canada | Patients with symptomatic systolic HF (i.e., with reduced EF) | 1-, 2-, and 3-year risk of all-cause mortality | Diuretic dose, SBP, lymphocyte, haemoglobin, uric acid, allopurinol use, LVEF, ischaemic aetiology, sodium, NYHA class | 2,987; international [15] |
| | | | | | | | 148; USA [15] |
| | | | | | | | 925; USA and Canada [15] |
| | | | | | | | 5,010; international [15] |
| | | | | | | | 872; Italy [15] |
| | | | | | | | 10,930; USA[16] |
| | | | | | | | 6,161; international[17] |
| Pocock et al. (2013) [7] | Meta-analysis Global Group in Chronic Heart Failure Model | 39,372 | International (30 cohorts) | Patients with diagnostic HF (i.e., both reduced and preserved EF) | 1-, and 3-year risk of all-cause mortality | Age, EF, NYHA class, creatinine, diabetes, beta-blocker use, SBP, BMI, HF duration, current smoker, COPD, gender, ACE-I or ARB use | 10,930; USA[16] |
| | | | | | | | 6,161; international[17] |
| | | | | | | | 39,372; internal [7] |
| | | | | | | | 6,263; international [10] |
| | | | | | | | 5,625; Korea [18] |
| | | | | | | | 51,043; Sweden [19] |
| McDowell et al. (2024) [10] | Prognostic Models for Mortality and Morbidity in Heart Failure With Preserved Ejection Fraction Models | 6,263 | International | Patients with symptomatic HF (i.e., both reduced and preserved EF) | 1-, and 2-year risk of cardiovascular death or HFH; cardiovascular death; all-cause mortality | NT-proBNP level, HFH within the past 6 months, creatinine level, diabetes, geographic region, HF duration, treatment with a sodium-glucose cotransporter 2 inhibitor, chronic obstructive pulmonary disease, transient ischemic attack/stroke, any previous HFH, and heart rate | 4,796; international [10] |
| | | | | | | | 4,128; international [10] |

ACE-I: angiotensin-converting enzyme inhibitor; ARB: angiotensin-receptor blockers; AUC: area under the curve; BMI: body mass index; COPD: chronic obstructive pulmonary disease; EF: ejection fraction; HFH: heart failure hospitalization; HF: heart failure; LVEF: Left ventricular ejection fraction; NT-proBNP: N-terminal pro–brain natriuretic peptide; NYHA: New York Heart Association; ROC: receiver operating characteristic; SBP: systolic blood pressure; NA, not applicable;